\documentclass[conference,letterpaper]{IEEEtran}

\usepackage[utf8]{inputenc} 
\usepackage[T1]{fontenc}
\usepackage{url}
\usepackage{ifthen}
\usepackage{cite}
\usepackage{hyperref}
\usepackage[cmex10]{amsmath} 
\usepackage[usenames,dvipsnames]{xcolor} 

\usepackage{amsfonts}       
\usepackage{nicefrac}       
\usepackage{microtype}      

\usepackage{commath}

\usepackage{algorithm}
\usepackage{algorithmic}
\usepackage{amsthm}
\usepackage{amssymb}
\usepackage{bbm}
\usepackage{paralist}
\usepackage{enumitem}
\usepackage{mathtools}
\usepackage{dsfont}
\usepackage{verbatim}
\usepackage{bm}

\newcommand{\lip}{\textnormal{lip}}

\newcommand{\KL}{\textnormal{KL}}
\newcommand{\kl}{\KL}
\newcommand{\TN}{\textnormal}
\newcommand{\defeq}{:=}

\newtheorem{theorem}{Theorem}

\newtheorem{lemma}{Lemma}
\newtheorem{remark}{Remark} 
\newtheorem{proposition}{Proposition}
\newtheorem{property}{Property}

\newtheorem{assumption}{Assumption}
\newcommand{\EXP}{\mathbb{E}}
\renewcommand{\Pr}{\mathbb{P}}
\renewcommand{\set}[1]{\mathcal{#1}}
\newcommand{\ep}{\hfill $\Box$}
\DeclareMathOperator*{\argmax}{arg\,max}
\DeclareMathOperator*{\argmin}{arg\,min}

\newcommand{\tred}[1]{{\color{red}#1}}
\renewcommand{\vec}[1]{\boldsymbol{#1}}

\usepackage{graphicx}
\graphicspath{ {./graphs/} }
\usepackage{subfigure}

\usepackage{xspace}
\usepackage{xcolor}
\DeclareMathOperator{\one}{\mathds{1}}
\newcommand{\eps}{{\varepsilon}}
\newcommand{\kj}[1]{{\color{RedOrange}KJ: #1}}

\usepackage{enumitem}

\def\cd{\cdot}
\newcommand{\blue}[1]{{\color[rgb]{0,0,0}#1}}
\def\eps{\ensuremath{\varepsilon}\xspace}
\def\hyphen{{\text{-\hspace{-.06em}}}}
\newcommand\kmax[1]{\mathop{#1\hyphen\max}}

\def\ddefloop#1{\ifx\ddefloop#1\else\ddef{#1}\expandafter\ddefloop\fi}
\def\ddef#1{\expandafter\def\csname c#1\endcsname{\ensuremath{\mathcal{#1}}}}
\ddefloop ABCDEFGHIJKLMNOPQRSTUVWXYZ\ddefloop
\DeclareMathOperator{\PP}{\mathbb{P}}
\newcommand{\sr}[2]{ {\stackrel{#1}{#2}} }
\newcommand{\fr}[2]{ { \frac{#1}{#2} }}
\def\lt{\left}
\def\rt{\right}
\def\bmu{{{\boldsymbol \mu}}}
\def\bx{{{\boldsymbol x}}}
\def\RR{{{\mathbb{R}}}}

\def\gam{{\gamma}}
\def\vmin{v_{\min}}

\newcommand{\hjcomment}[1]{{\textbf{{\textcolor{blue}{HJ --- #1}}}}}

\begin{document}
\title{Transfer Learning in Bandits with Latent Continuity} 



\author{%
  \IEEEauthorblockN{Hyejin Park\IEEEauthorrefmark{2}\IEEEauthorrefmark{1},
                    Seiyun Shin\IEEEauthorrefmark{3}\IEEEauthorrefmark{1},
                    Kwang-Sung Jun\IEEEauthorrefmark{4},
                    and Jungseul Ok\IEEEauthorrefmark{2}}
  \IEEEauthorblockA{\IEEEauthorrefmark{2}%
                    POSTECH,
                    \{parkebbi2, jungseul\}@postech.ac.kr}
  \IEEEauthorblockA{\IEEEauthorrefmark{3}%
                    University of Illinois at Urbana-Champaign,
                    seiyuns2@illinois.edu}
  \IEEEauthorblockA{\IEEEauthorrefmark{4}%
                    University of Arizona,
                    kjun@cs.arizona.edu}
}


\maketitle
\def\thefootnote{*}\footnotetext{Authors contributed equally.}



\begin{abstract}
Structured stochastic multi-armed bandits provide accelerated regret rates over the standard unstructured bandit problems.
Most structured bandits, however, assume the knowledge of the structural parameter such as Lipschitz continuity, which is often not available.
To cope with the latent structural parameter, we consider a transfer learning setting in which an agent must learn to \textit{transfer} the structural information from the prior tasks to the next task, which is inspired by practical problems such as rate adaptation in wireless link.
Specifically, we propose a simple but efficient framework to provably and accurately estimate the Lipschitz constant based on previous tasks and fully exploit it for the new task at hand. We analyze the efficiency of the proposed framework in two folds:
(i) our regret bound on the new task is close to that of the oracle algorithm with the full knowledge of the Lipschitz constant under mild assumptions; and (ii) the sample complexity of our estimator matches with the information-theoretic fundamental limit.
Our analysis reveals a set of useful insights on transfer learning for latent Lipschitz constants such as the fundamental challenge a learner faces.  
Finally, our numerical evaluations confirm our theoretical findings and show the superiority of the proposed framework compared to baselines.

\end{abstract}





\section{Introduction}

The classical stochastic multi-armed bandit (MAB) 
\cite{Robbins1952} 
of independent $K$ arms in time-horizon $T$
has
the fundamental limit of regret
$O(K \log T)$
which scales linearly with the number of arms $K$ and thus not practical when $K$ is very large.
To deal with large number of arms, 
one may exploit known correlations among the arms via a structural assumption.
This idea has resulted in a stream
of research activities on bandits under a variety of structural assumptions, e.g.,
Lipschitz \cite{Magureanu2014Lipschitz}, 
linear \cite{dani2008stochastic}, convex \cite{agarwal2011stochastic}, or unimodal \cite{Yu2011Unimodal}.
These studies
are meaningful in deriving bandit algorithms whose regret is \textit{scale-free}, indicating
that it does not grow with $K$ for large enough $K$.
In case of
the Lipschitz continuity structure described by a Lipschitz constant $L$~\cite{ok2018exploration},
for example, one can achieve $O(\min \{K, \text{poly}(L)\}\log T)$.


Such a benefit from the continuity structure, however, requires prior knowledge on the Lipschitz constant $L$,
which is often latent in practice.
In this case, it is natural to estimate the latent $L$ from similar tasks.
For example, rate adaptation in wireless link \cite{combes2018optimal, qi2019rate} faces a sequence of bandit problems with similar structural properties,
while one can find a continuity structure among throughputs of transmission rates.
The channel often changes discretely over time, and these changes partition a sequence of tasks as in our learning scenarios.
This motivates us to study a \textit{transfer learning} problem
where we aim to learn the latent Lipschitz constant $L$
from previous tasks and use it for the next task.

In order to build provably sample-efficient algorithms for the transfer bandit problem, we first investigate 
the risks of using wrong estimation of Lipschitz constant $L$
in two main failure scenarios (Section~\ref{sec:L-impact}).
Overestimating $L$ leads to an unnecessary regret, attenuating the benefit from the structure such as scale-free regret.
On the other hand, underestimating $L$ can cause a catastrophic failure of suffering a linear regret.
We design an estimator for $L$, which balances between the two extremes.
We show that 
using this estimator, one can nearly achieve the minimal regret of the oracle algorithm knowing $L$ exactly 
(Section~\ref{sec:proposed-algorithm}). 
Furthermore, our estimator
is asymptotically optimal in 
that the sample complexity on previous tasks matches to the one in a fundamental limit analysis. 
Finally, we conclude with numerical verification of our theoretical findings (Section~\ref{sec:numerical}) with exciting future directions.

\smallskip
\noindent{\bf Related work.}
In \cite{Combes2017OSSB}, the authors provide
a generic approach to construct optimal algorithms
when we have the complete knowledge on 
the structural property, including but not limited to
Lipschitz \cite{Magureanu2014Lipschitz}, 
linear \cite{dani2008stochastic}, convex \cite{agarwal2011stochastic}, or unimodal \cite{Yu2011Unimodal}.
The structural knowledge, however, 
is often incomplete in practice as mentioned above.
For this issue,
in \cite{bubeck2011lipschitz},
it is shown to be possible to achieve
the minimax optimal regret of 
$\Theta(L^{D/(D+2)} T^{(D+1)/(D+2)})$ 
without any knowledge of the Lipschitz constant for the continuum of arms, where $D$ is the dimension of the embedding space.
The minimax optimality compares the regret upper bound of the algorithm to the worst lower bound of regret.
Although such a minimax analysis is inevitable for the continuum of arms, a minimax optimal algorithm can be very far from the 
instant-dependent optimality particularly in canonical cases.
Thus, although achieving both minimax and instance-dependent optimality is an active area of research~\cite{tirinzoni2020novel}, we focus on the instance-dependent regret bounds.

The pioneering work of 
\cite{Emma2013SequentialTransfer} has proposed a transfer learning framework where
the learner faces a sequence of bandit tasks that are randomly drawn from a distribution over a \textit{finite} set of problem instances.
They propose an algorithm that leverages the robust tensor power method to learn the underlying set of 
instances while repeatedly solving each task.
Their algorithms, however, only consider a finite number of instances, in contrast to our Lipschitz structure which consists of infinitely many instances.
While there have been a few follow-up studies for infinite instance sets such as~\cite{Lazaric2014MultiLinBandits, Lazaric2020MultiLinBandits}, 
they both consider the simple linear structure and focus on minimax regret rather than instance-dependent regret.
To our knowledge, we are the first to study transfer learning in bandits with instance-dependent optimality beyond the simple case of the finite instance set. 
\section{Preliminary}

\subsection{Lipschitz bandit model
}
Let $\vec{\mu}= (\mu(1), ..., \mu(K))$ denote a multi-armed bandit instance, where
each play of arm $i \in [K]:= \{1, 2, ..., K \}$ generates a reward drawn i.i.d. from Bernoulli distribution with mean $\mu(i) \in [0,1]$.
Notice that the choice of reward distribution and our analysis can be generalized to those related to exponential family with a single parameter~\cite{garivier2011kl,kaufmann2016bayesian}, but for ease of exposition, we restrict our attention to Bernoulli rewards. 
At each round $t = 1,2, ...,$ the decision maker $\pi$ selects an arm $i_t$, pulls it, and then receives a reward $r_t$ drawn from the distribution associated with the arm $i_t$.
Let $\mu_* := \max_{i \in [K]} \mu(i)$
and $\set{K}_*(\vec{\mu}) := \{i \in [K] : \mu(i) = \mu_*\}$
denote the best mean reward and the set of best arms, respectively.
%
For a given MAB instance $\vec{\mu}$, an algorithm $\pi$ aims to maximize the expected cumulative rewards over the time horizon $T$.
This aim is equivalent to minimizing the regret defined as follows:
\begin{align*}
R^\pi_T(\vec{\mu}) \defeq \sum_{i \in [K]}( \mu_*  - \mu(i)) \EXP_\pi [n_T (i) ] \;,
\end{align*}
where $n_T(i) $ is the number of pulling arm $i$ up to time $T$,
and the expectation $\EXP_\pi$ is taken w.r.t. the randomness induced by both the rewards and the algorithm~$\pi$.
The regret $R^\pi_T(\vec{\mu})$ can also be viewed as the expected opportunity cost for selecting sub-optimal arms.




\subsection{Optimal regret with known $L$}
We consider the set of mean rewards where the arms are constrained to satisfy Lipschitz condition w.r.t. an embedding of the arms $\vec{x} = (x(1), ..., x(K)) \in [0,1]^{D \times K}$, which is commonly referred to as the Lipschitz structure \cite{Magureanu2014Lipschitz}.
Specifically, the Lipschitz structure $\Phi(L)$ for a given Lipschitz constant $L>0$ is defined as follows: 
\begin{align*}
\!\Phi(L)
\defeq  \left\{\vec{\mu}  \!\in \! [0,1]^K\!\!:  |\mu(i) \!-\! \mu(j)| \le L  d(i,j) ~ \forall i, j\!\in \![K] \right\} \;,
\end{align*}
where $d(i,j) \defeq \|x(i) - x(j)\|$.
We assume that the learner knows that the instance $\vec{\mu}$ conforms to the structure $\Phi(L)$. 
We say an algorithm~$\pi$ is uniformly good for $\Phi(L)$
if  $\EXP_\pi [n_T (i)] = o(T^\rho)$ for all $i \notin \set{K}_*(\vec{\mu})$, 
$\vec{\mu} \in \Phi(L)$, and $ \rho > 0$.
That is, a uniformly good algorithm
has ability to adapt to any $\vec{\mu}  \in \Phi(L)$ and enjoys a sublinear
regret in $T$.
Then, uniformly good algorithms have the following fundamental limit:
\begin{theorem}[Regret lower bound with known $L$~\cite{Magureanu2014Lipschitz}] \label{thm:lower}
Let $\pi$ be a uniformly good algorithm for $\Phi(L)$.
For any $\vec{\mu} \in \Phi (L)$,
\begin{equation}\label{eq:low1}
\liminf_{T\to\infty} \frac{R^\pi_T(\vec{\mu})}{\log T} \ge C(\vec{\mu}, L) \;,
\end{equation}
where $C(\vec{\mu}, L)$ is the optimal value of the following linear programming (LP):
\begin{subequations}
\label{eq:opt}
\begin{align}
    \label{eq:objective}
    \underset{
    \vec{\eta} \succcurlyeq 0 
    }{\TN{min}} & ~~  \sum_{i \notin \set{K}_* (\vec{\mu})} \!\! \big(\mu_* - \mu(i) \big) \eta(i)   \\
      \TN{s.t.~}
    & \!\sum_{i \notin \set{K}_* (\vec{\mu})} \!\!\! \kl(\mu(i) \| \nu^{j}(i ; \vec{\mu}, L))  \eta(i) \ge 1 
    ,\forall j \notin\set{K}_* (\vec{\mu}).
    \label{eq:feasible2}
\end{align}
\end{subequations}
%
Here 
$\nu^{j}(i ; \vec{\mu}, L) \defeq \max \{\mu(i), \mu_* - L d(i,j) \}$ for all $i,j \in [K]$,
and $\kl(\mu \| \nu)$ is the Kullback-Leibler divergence 
between Bernoulli distributions with mean $\mu$ and $\nu$, i.e.,
$\kl({\mu} \| {\nu}) \defeq \mu \log \left( \mu/\nu\right)
+(1-\mu) \log ((1-\mu)/(1-\nu))$.

%
%
\end{theorem}

\noindent{\bf Optimal algorithms for known $L$:}  \label{model:known_structure_opt_alg}
There exist a number of algorithms that provably achieve
the fundamental limit in Theorem~\ref{thm:lower} such as
\cite{Magureanu2014Lipschitz, Combes2017OSSB, ok2018exploration}.
Briefly, at each iteration $t$, the algorithm 
finds a solution of LP in \eqref{eq:opt}, denoted by $\vec{\eta}(\vec{\hat{\mu}_t}, L)$,
based on estimated parameter $\vec{\hat{\mu}}_t$
and 
tracks the solution in a suggestion on minimal exploration so that 
${\eta}(i; \vec{\hat{\mu}_t}, L) \le \frac{n_t(i)}{\log t}$ for each suboptimal arm $i$.
We use
a simplified version of algorithm proposed in \cite{ok2018exploration}, denoted by $\pi(L)$, 
which
is originally designed for structured Markov decision process.
Due to the space limit, we postpone the detailed algorithm to the full version.
For any $\vec{\mu} \in \Phi(L)$, the algorithm $\pi(L)$ enjoys the following regret upper bound. 

\textfloatsep=.5em

\begin{theorem}[Regret upper bound with known $L$~\cite{Magureanu2014Lipschitz}] \label{thm:upper} 
Consider $\vec{\mu} \in \Phi (L)$ such that
for each $i \notin
\set{K}_{*} (\vec{\mu})$, 
the LP solution 
$\eta(i; \vec{\mu}, L)$ is unique and continuous at $\vec{\mu}$. 
Then, for any given $\lambda >0$, algorithm~$\pi = \pi(L)$ has
\begin{equation}\label{eq:upper}
\limsup_{T\to\infty} \frac{R^{\pi}_T(\vec{\mu})}{\log T} \le 
(1+\lambda) C(\vec{\mu}, L) \;.
\end{equation}
\end{theorem}
Theorem~\ref{thm:upper} implies that 
when using exact value of $L$, we can asymptotically achieve the regret lower bound in 
Theorem~\ref{thm:lower}.
It is worth noting that exploiting the exact Lipschitz structure provides a drastic reduction in
regret:

\begin{property} \label{prop:C}
Let $\Delta_{\vec{\mu}} \defeq \min_{i \notin \set{K}_* (\vec{\mu})}  \mu_*  - \mu(i)$
denote the smallest suboptimality gap. Then,
for $L > 0$ and $\vec{\mu} \in \Phi(L)$, 
\begin{align} \label{eq:C-bound}
C(\vec{\mu}, L) \le
\frac{8}{\Delta^2_{\vec{\mu}}}\min \left\{ 
K, 
\left( 
\frac{8 L \sqrt{D}}{\Delta_{\vec{\mu}}}
+1
\right)^D
\right\}    \;.
\end{align}
\end{property}

Given a fixed $\Delta_{\vec{\mu}}$ and $D$, 
the fundamental limit of regret $C(\vec{\mu}, L)$
is scale-free, meaning that it does not scale with the number of arms $K$.
In particular, this is a dramatic advantage over 
no continuity structure, i.e., $L=\infty$,
whose optimal regret does scale with $K$. 

\begin{figure}
\vspace{-0.5cm}
    \centering
    \subfigure[Bandit parameter $\vec{\mu}$ and embedding $\vec{x}$ with $L = 200$]{
        \includegraphics[width=0.38\textwidth, trim={0 0.1cm 0 1cm}, clip]{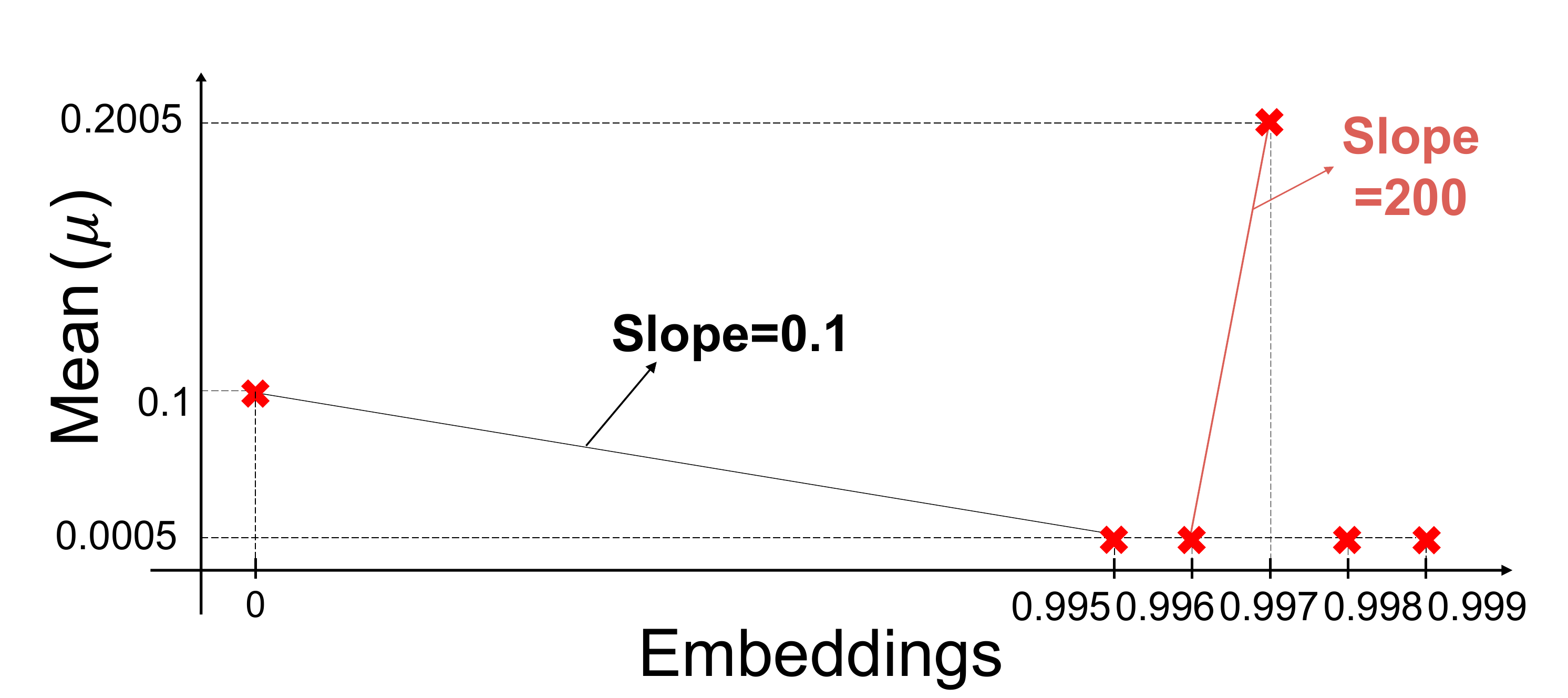}
        \label{fig:risk:para}}

    \subfigure[Regret over time]{
        \includegraphics[width=0.38\textwidth, trim={0 0.2cm 0 1.6cm}, clip]{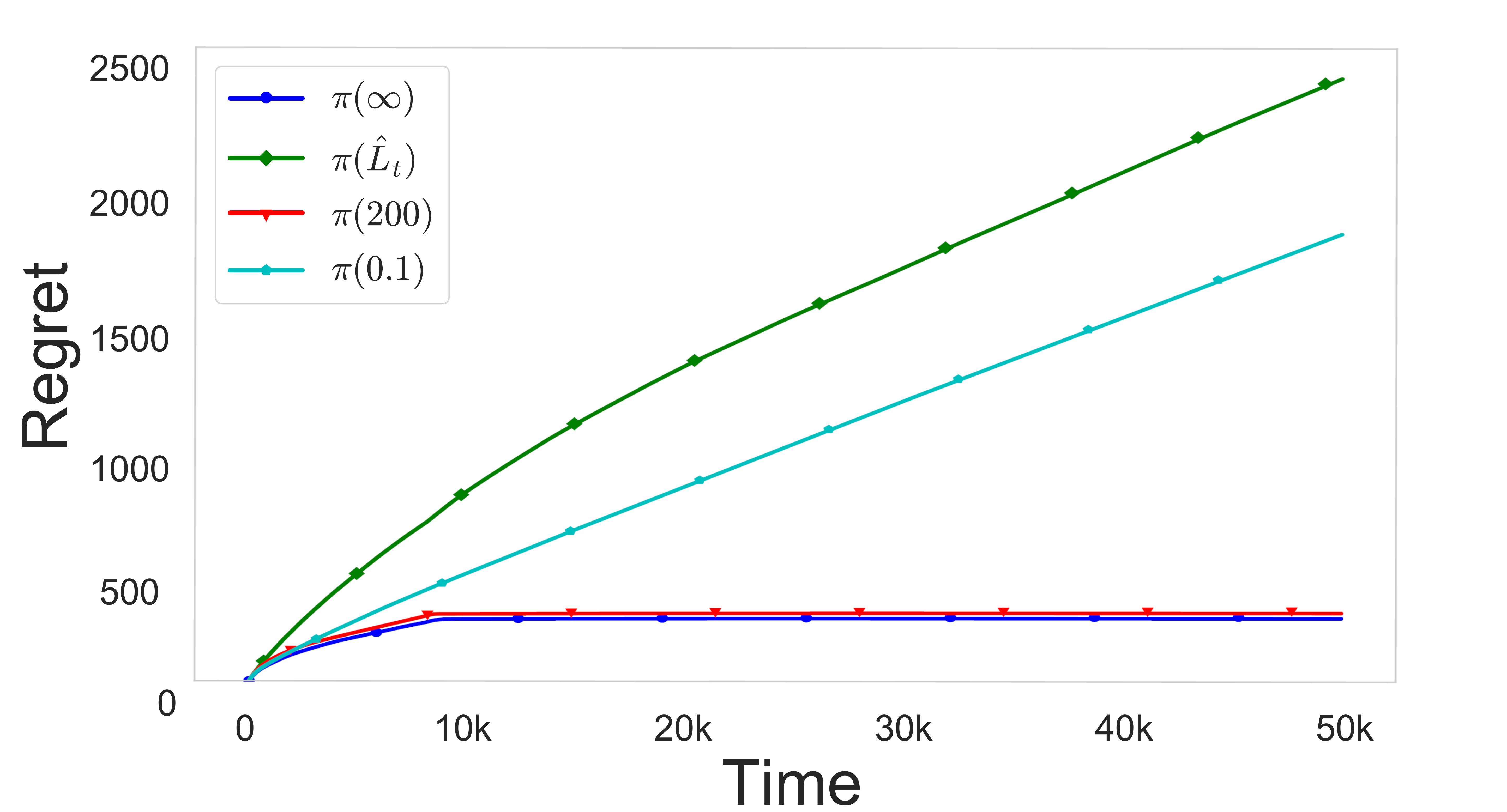}
        \label{fig:risk:regret}}
    \vspace{-0.6em}
    \caption{Comparison among $\pi(\infty)$, $\pi(200)$, $\pi(0.1)$ and $\pi(\hat{L}_{t})$
    for given $\vec{\mu}$ and $\vec{x}$ shown in Figure~\ref{fig:risk:para}.} 
    \label{fig:risk}

\end{figure}

\subsection{Impact of incorrect estimation of $L$}
\label{sec:L-impact}

In the context of \emph{latent} Lipschitz constant $L$, 
one needs to estimate $L$ from observed samples, which can be quite inaccurate.
We study the impact of using an incorrect estimator $L'$ in the following two cases: (i) $L' > L$; and (ii) $L'< L$. 

\smallskip
(i) $L' > L$:
In this case, the regret of algorithm~$\pi (L')$ is provably bounded from above $C(\vec{\mu}, L')$, and thus causes a larger regret. This follows from the fact that for any $L' \ge L$, $\Phi(L) \subset \Phi(L')$, so we have $C(\vec{\mu}, L) \le C(\vec{\mu}, L')$, $\forall \vec{\mu} \in \Phi(L)$.
The regret rate of algorithm $\pi(L')$ can be degenerated into that of the unstructured case due to the conservative choice of $L'$. Then,
the regret must scale with $K$ as discussed above, which is problematic for larger $K$'s.

\smallskip
(ii) $L'< L$: In this case, $L'$ is underestimated, which implies that $\vec{\mu} \in \Phi (L) \setminus \Phi(L') \neq \emptyset$. 
Therefore, the regret bound stated above does not hold anymore.
The algorithm $\pi(L')$ may suffer a linear regret
because once the algorithm starts exploiting an incorrect best arm, it may not be able to collect sufficient statistical evidence to correct itself.

\smallskip
We confirm this phenomenon by a numerical simulation in Figure 1, where the mean rewards of 6 arms $\vec{\mu}=(0.1, 0.0005, 0.0005, 0.2005, 0.0005, 0.0005)$ and their embedding $\vec{x}=(0, 0.995, 0.996, 0.997, 0.998, 0.999)$ are illustrated in Figure~\ref{fig:risk:para}.
We set the true Lipschitz constant $L$ to be $200$ and the second steepest slope to be $0.1$.
For each time step $t$, we define the estimator $\hat{L}_t \defeq \max_{i \neq  j \in [K]} \frac{|\hat{\mu}_t (i)- \hat{\mu}_t (j)|}{d(i,j)}$.
With the time horizon $T=50,000$, Figure~\ref{fig:risk:regret} compare four algorithms:
$(a)\ \pi(\infty)$; $(b)\ \pi(\hat{L}_{t})$; $(c)\ \pi(200)$; and $(d)\ \pi(0.1)$.
We remark that $\pi(\hat{L}_t)$ updates the estimator after each time step.
We observe that the most conservative choice of $L = \infty$ and the exact choice of $L = 200$
    show similar logarithmic regrets.
On the other hand, the aggressive choice of $L = 0.1$ suffers from an almost linear regret due to the mismatch between the true structure and its belief.
The last method $\pi(\hat{L}_t)$ also 
shows an almost linear regret and even worse regret than $\pi(0.1)$, despite keeping to update its estimator for $L$.
This is because the Lipschitz constant is underestimated by undersampling the hidden true best arm~$4$ (i.e., $x=0.997$).
This underestimated $\hat{L}_t$ reinforces the algorithm to reduce the exploration rate on the true best arm, thus losing the opportunity to correct $\hat{L}_t$.

To summarize, our simulation shows that simultaneously learning structure and minimizing regret in $\pi(\hat{L}_t)$ in transfer learning setting is highly nontrivial and cannot be done via naive methods.

\section{Main Results}
\label{sec:unknown}


\subsection{Transfer learning model}

\begin{figure}[!t]
\centering
\vspace{-0.6cm}
\includegraphics[width=0.38\textwidth, trim={0 0.1cm 0 0.2cm}, clip]{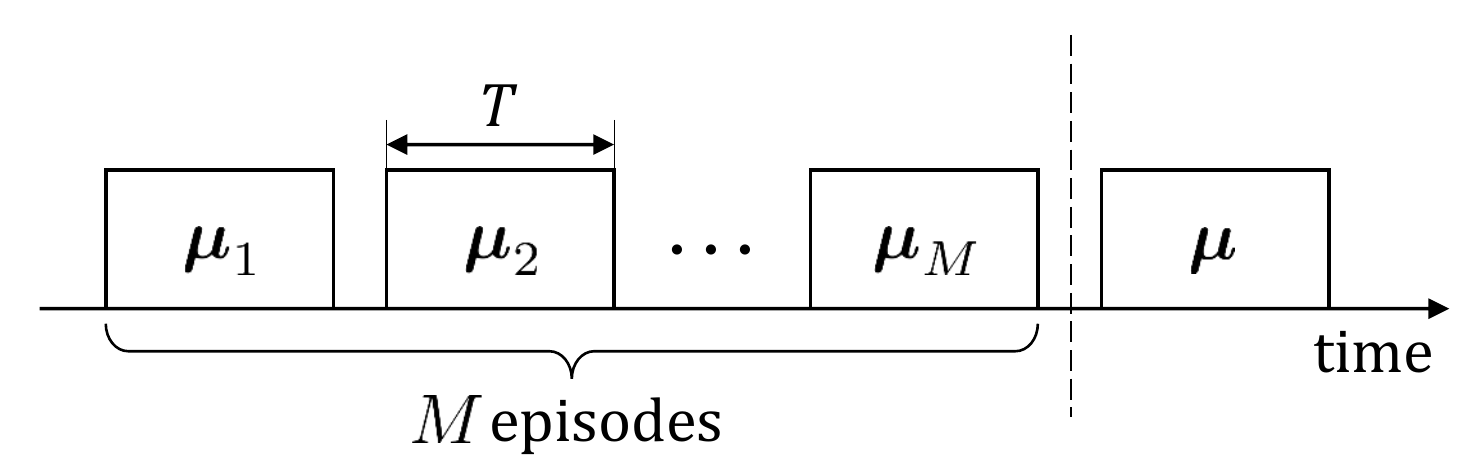}
\vspace{-.5em}
\caption{Transfer learning from previous $M$ episodes. \label{fig:framework}}
\vspace{-.3em}
\end{figure}

Hereafter, we assume that the Lipschitz constant $L > 0$ is latent, although we know the embedding $\vec{x}$ in advance, i.e.,
the relative similarity $(d(i, j))_{i, j \in [K]}$ among the arms.
To learn latent $L$, we consider a scenario of transfer learning illustrated in Figure~\ref{fig:framework} where
one aims to transfer the knowledge on $L$ extracted from $M$ past episodes with the mean rewards $(\vec{\mu}_m)_{m \in [M]}$
satisfying $\vec{\mu}_m \in \Phi(L)$
to a new episode $(M+1)$ with mean rewards $\vec{\mu} \in \Phi(L)$. For simplicity, we assume that each episode has the same length $T$.
Let $L_m \defeq \max_{i \neq j \in [K]} \frac{|\mu_m(i)  - \mu_m(j)|}{\| x(i) - x(j)\|}$
be the tightest Lipschitz constant of episode $m$.
Note that $\max_{m \in [M]} L_m \le L$.
We make the following assumption on $L_m$'s:
\begin{assumption}[Learnability]  \label{asm:tight}
At least $\alpha$-portion of the previous $M$ episodes have their $L_m$ close to $L$ with certain margin 
$\varepsilon_\alpha  >0$. 
Formally, there exist $\alpha >0$ and $\varepsilon_\alpha > 0$ such that
\begin{align*}
|\{m\in[M]: L_m \ge L - \eps_\alpha\} | \ge \alpha M  \;.
\end{align*}
\end{assumption}
The parameters $\alpha$ and $\eps_\alpha$ in Assumption~\ref{asm:tight} quantify the difficulty of estimating $L$ tightly, 
where larger $\alpha$ and smaller $\eps_\alpha$
imply sharper concentration of $L_m$'s 
around $L$ and thus easier setting.
Recalling that smaller $L$ implies smaller 
regret, 
we aim to estimate the smallest possible $L$.
Notice that 
$L = \max_{m \in [M]} L_m$ can be easily manipulated with
any large $L_m$.
We can assume a prior distribution of bandit parameter over structure $\Phi(L)$ instead of Assumption~\ref{asm:tight},
but it is somewhat impractical to know this.
Thus, through the assumption on tail distribution of $L_m$’s which allows a robust estimation of $L$, it is possible to 
securely recover 
the Lipschitz constant. 
In addition,
our analysis under Assumption~\ref{asm:tight} can be easily applied to 
the case assuming prior distribution of bandit parameter.
%
The difficulty of estimating $L$ also depends on 
the sampling scheme in the prior tasks.
Hence, we make the following assumption:
\begin{assumption}[Minimal exploration] \label{asm:minimal}
For each episode $m$, every arm $i \in [K]$ is pulled at least
 $\tau > 0$.
\end{assumption}
In Assumption~\ref{asm:minimal}, a small value of $\tau$
implies a high risk of having insufficient samples
for estimation of $L$.




\subsection{Extracting Lipschitz constant}
\label{sec:proposed-algorithm}


To analyze with latent $L$, 
we take a two-folded approach. First, we estimate $L$ from extracting structural information in previous $M$ episodes. We then run $\pi(\hat{L}_\beta)$, where $\hat{L}_\beta$ is the estimate $L$.
Let $\hat{\vec{\mu}}_m$ be the estimated 
mean rewards
and ${\hat{L}_m} \defeq \max_{i \neq j \in [K]} \frac{|\hat{\mu}_{m}(i) - \hat{\mu}_{m} (j)|}{\|x(i) - x(j)\|}$ be the estimated Lipschitz constant in episode $m$.
For an efficient estimate of $L$, we introduce two hyperparameters $\beta \in (0, \alpha)$ and $\eps_\beta > \eps_\alpha$. We set our estimator as an upper confidence bound on $L$:
\begin{align}\label{eq:typical-L}
  \hat{L}_\beta  \defeq \ell_\beta+\varepsilon_\beta.
\end{align}
Here $\ell_\beta \defeq \kmax{\lceil\beta M\rceil}_{m\in[M]} ~\hat L_m$, where $\kmax{k}$ denotes the operator taking the $k$-th largest element.
Note that the margin $\eps_\beta$ is imposed to reduce the risk of underestimating $L$
since
structured bandit algorithms with 
the underestimated Lipschitz constant
can induce almost linear regret 
as shown in Section~\ref{sec:L-impact}.
By running $\pi(\hat{L}_\beta)$, we get the following performance guarantee:

\begin{theorem} \label{thm:transfer}
Suppose Assumptions~\ref{asm:tight}~and~\ref{asm:minimal} hold for $\alpha > 0$ and $\varepsilon_\alpha >0$.
Let $\beta \in (0,\alpha)$ and 
$\tau \ge \fr{4}{\Delta^2_\bx (\eps_\beta-\eps_\alpha)^2  }
\left(
 \ln(2K) + \frac{1}{\min \{ \beta, \alpha - \beta\}} 
\right)$.
If $M \ge 2Z \ln(2ZT)$ with
 $Z=\fr{1}{\min\cbr{\beta,\alpha-\beta}  \ln(2K)}$,
for any $\bmu \in \Phi(L)$, algorithm~$\pi(\hat{L}_\beta)$ with $\lambda  > 0$
has
\begin{align*}
 & \limsup_{T \to \infty} \frac{R^\pi_T(\vec{\mu})}{ \log T} \le 
  (1+\lambda) \ C\!\left(\vec{\mu}, L+2\varepsilon_\beta - \varepsilon_\alpha \right).
\end{align*}
\end{theorem}
\begin{proof}
See Section~\ref{sec:pf-transfer}.
\end{proof}

The upper bound of
$C(\vec{\mu}, L')$  in 
\eqref{eq:C-bound} is continuous  $L'$
for $L' \ge L$ and $\vec{\mu} \in \Phi (L)$.
Besides,
under mild additional assumptions, we also have a continuity of $C(\vec{\mu}, L')$  
in $L'$ for $L'\in [L, L +\varepsilon)$
and $\vec{\mu} \in \Phi (L)$.
We provide a formal description of continuity of $C(\vec{\mu}, L)$ in $L$, implying a near optimality of the regret upper bound of $\pi(L_\beta)$.

\begin{theorem} \label{thm:continuity}
For given bandit structure $\Phi(L')$ with fixed $\vec{\mu}$ and $\vec{x}$, the optimal value of \eqref{eq:objective}\textendash\eqref{eq:feasible2}, $L' \rightarrow C(\vec{\mu}, L')$ is continuous in $[L, L +\delta)$,
provided that the optimal arm $x^*(\vec{\mu})$, the solution to problem \eqref{eq:objective}\textendash\eqref{eq:feasible2} in Theorem~\ref{thm:lower}, are unique and  $\delta$ satisfies
\begin{align} \label{eq:delta_cond}
    0 \leq \delta < \min_{i, j \in [K]}\left\{\frac{\mu_{*}-\mu(i)}{\|x(i) - x(j)\|}-L\right\} \;,
\end{align}
such that 
$\mu(i) > \sqrt{DK}\delta$ and
$\nu^j(i; \vec{\mu}, L) < 1-\sqrt{DK}\delta$ for all $i, j \in [K]$.
\end{theorem}
Hence, Theorem~\ref{thm:transfer}
implies that 
when $\tau$ and $M$ are sufficiently large, i.e.,
we have rich experiences with prior tasks, 
the algorithm~$\pi(\hat{L}_\beta)$
closely achieves the fundamental limit
of oracle performance knowing $L$ in advance.
One can interpret $\tau M$ as the sample complexity
for some probably approximately correct (PAC) 
learning of $L$.
The sample complexity required in
Theorem~\ref{thm:transfer} is
\begin{align} \label{eq:sample-required}
         \tau M = \Omega \left(
     \frac{1}{
     {\Delta_\bx^2 (\varepsilon_\beta - \varepsilon_\alpha)^2} \alpha}
     \log T \right) \;, 
\end{align}
which matches with the information-theoretic lower bound
for the PAC learning obtained in 
Section~\ref{sec:pf-transfer}.



\begin{remark}[Hyperparameter choice for $\hat{L}_\beta$] \label{sec:hyper}
Theorem~\ref{thm:transfer} includes an intrinsic trade-off for the choice of $\beta$, which appears in the requirement of $M \ge \tilde\Theta(\max\{\fr1\beta, \fr1{\alpha-\beta}\})$. Here $\tilde\Theta$ hides logarithmic factors.
Notice that when $\beta$ is too small (choosing near the top of $\{\hat L_m\}$), the estimate $\hat{L}_\beta$ can be too large and overshoot, whereas when $\beta$ is too large, the estimate falls below $L$, incurring linear regret.

Our theorem also suggests that when a valid $(\alpha,\eps_\alpha)$ is available, one should set $\eps_\beta \approx 2\eps_\alpha$ and $\beta \approx \alpha/2$. It turns out that the algorithm is then guaranteed to use $\hat{L}_\beta$ that is at most $3\eps_\alpha$ away from $L$.
When the knowledge on $(\alpha,\eps_\alpha)$ is not available, one can see that $\eps_\beta$ should be greater than $1/M$. This is due to the fact that if we set $\beta=1/M$, then assuming all other variables are fixed, the requirement on $M$ becomes $M = \Omega( M \ln(M))$, which clearly cannot be satisfied for large enough $M$.
In other words, there is no guarantee that transfer learning will happen with our estimator $\hat{L}_\beta$.
This motivates us to adopt a quantile $\ell_\beta$ to compute $\hat{L}_\beta$,  rather than simply setting the estimate as the maximum of $L_m$'s.
\end{remark}

\subsection{Lower bound of sample complexity $\tau M$}
\label{sec:lower-bdd-sample-complexity}

We study a fundamental limit of sample complexity in estimating $L$ from $M$ prior tasks.
Let $\tau_{m}$ denote the number of playing 
the most under-sampled arm in episode~$m$.
Then, Assumption~\ref{asm:minimal} can be 
written equivalently as $\min_{m \in [M]} \tau_m \ge \tau$.
For given $\alpha >0, \varepsilon_\alpha > 0, \tau > 0$,
and $\varepsilon > \varepsilon_\alpha$,
we say an estimator $\hat{L}$
is uniformly good for $(\alpha, \eps_\alpha, \tau, \eps)$
if for any $L$,
$(\vec{\mu}_m)_{m\in [M]}$, and $(\tau_m)_{m \in [M]}$ such that $\vec{\mu}_m  \in \Phi (L)$ for each $m \in [M]$
satisfy Assumptions~\ref{asm:tight} and~\ref{asm:minimal},
then 
\begin{align} \label{eq:concen-L-required}
\Pr[ \hat{L} > L+ \varepsilon] +
\Pr[ \hat{L} < L]
\le o\left(\frac{\log T}{T} \right) \;.
\end{align}
The concentration in \eqref{eq:concen-L-required}
is a minimal condition to conclude
the desired regret upper bound in Theorem~\ref{thm:transfer}.
\begin{theorem}  \label{thm:sample-lower}
Suppose that an estimator $\hat{L}$
is uniformly good for $\alpha >0, \varepsilon_\alpha > 0, \tau > 0$, and $\varepsilon > \varepsilon_\alpha$.
Then, we must have
\begin{align} \label{eq:complexity-lower}
{\Delta_\bx^2 (\varepsilon - \varepsilon_\alpha)^2} \alpha \tau M = \Omega(\log T) \;.
\end{align}
\end{theorem}
Note that when we set $\eps = 2 \eps_\beta - \eps_\alpha$,
the concentration of $\hat{L}$ in \eqref{eq:concen-L-required}
becomes the one that we need to conclude
the regret bound of $\pi(\hat{L})$ in Theorem~\ref{thm:transfer}.
Hence, Theorem~\ref{thm:sample-lower} 
provides a lower bound on
$\tau M$ to obtain the desired concentration of $\hat{L}$
as~\eqref{eq:sample-required}
which asymptotically 
matches with the lower bound on $\tau M$ in Theorem~\ref{thm:upper} with $\beta = c \alpha$ for any positive constant $c < 1$
since the choice of $\beta$ implies
$ \min \{ \beta, \alpha - \beta\} = \min \{ c \alpha, (1-c) \alpha\}$.

\begin{figure*}[!ht]
    \centering
        \subfigure[Five incidences of $\vec{\mu}_m$]{
        \includegraphics[width=0.315\textwidth, trim={1.5cm 0.5cm 4.6cm 1cm},clip]{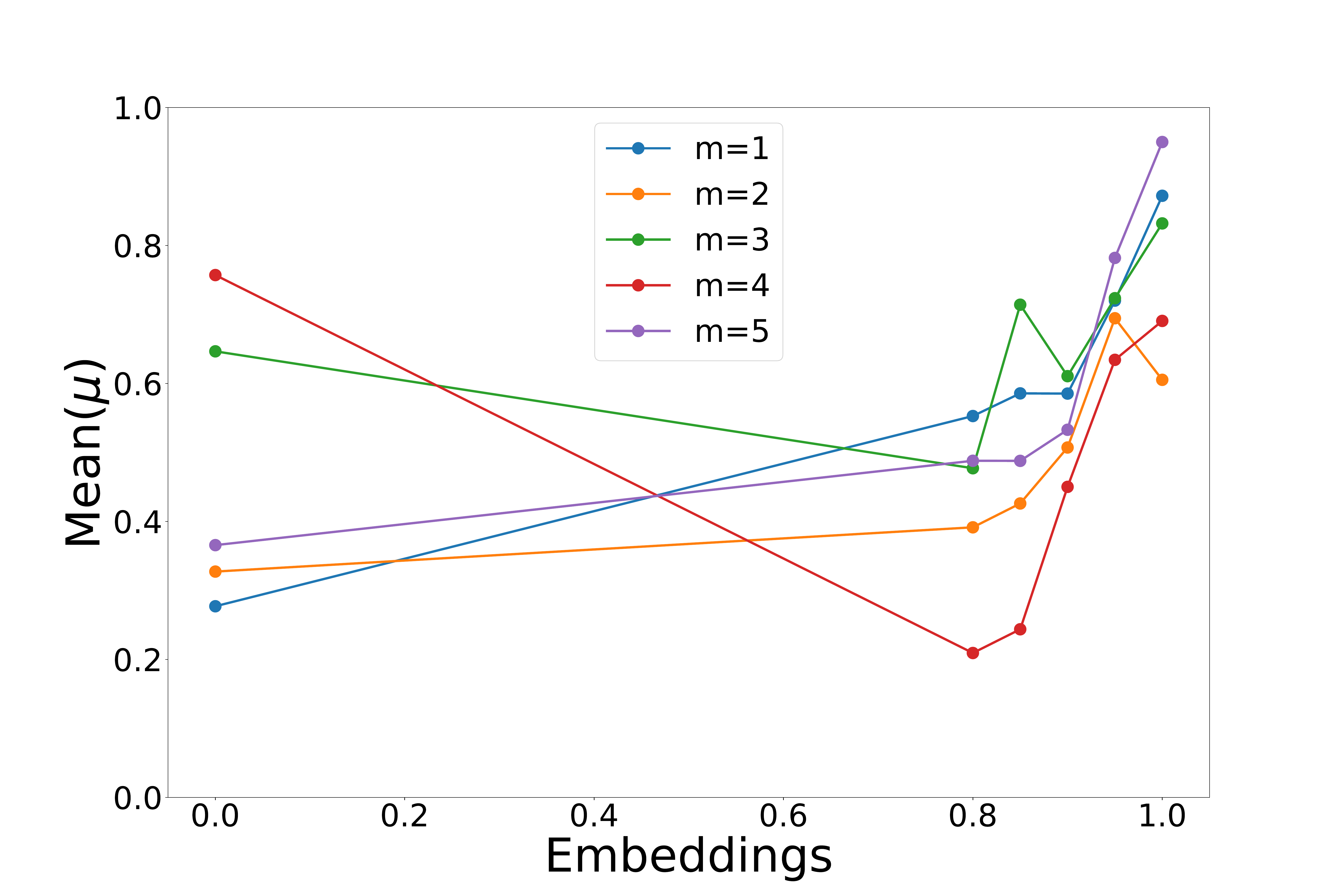}
        \label{fig:num:eg:L5}
        }
    \subfigure[Evolution of estimation over episodes]{
        \includegraphics[width=0.315\textwidth, trim={1.5cm 0.5cm 4.6cm 1cm},clip]{simul_Lbeta.pdf}
        \label{fig:num-re:estimate:L5}
        }
    \subfigure[Cumulative regret over episode]{
        \includegraphics[width=0.315\textwidth, trim={0cm 0.5cm 6.1cm 1cm},clip]{simul_cumregret5.pdf}
        \label{fig:num-re:regret:L5}
        }
    \vspace{-0.6em}
    \caption{Comparison of various estimators using the past experiences for $L = 5$.}
    \label{fig:regret:L}
    \vspace{-1.2em}
\end{figure*}




\subsection{Proof of Theorem~\ref{thm:transfer}} \label{sec:pf-transfer}
Let $\blue{\Delta_\bx} \defeq  \min_{i \neq j} \|d(i,j)\| > 0$.
The key analysis of Theorem~\ref{thm:transfer} lies in deriving the concentration of $\hat{L}_\beta$ to $L$:
\begin{lemma}\label{lem:concen}
  Take Assumptions~\ref{asm:tight}~and~\ref{asm:minimal} with $\alpha$ and $\varepsilon_\alpha$.
  Let $\beta \in (0, \alpha)$ and $\blue{\varepsilon'} :=  \varepsilon_\beta - \varepsilon_\alpha > 0$.
  If $\tau \ge \fr{4}{\Delta^2_\bx (\eps_\beta-\eps_\alpha)^2  }
\left(
 \ln(2K) + \frac{1}{\min \{ \beta, \alpha - \beta\}} 
\right)$,
  \begin{align} \label{eq:L-beta-concen}
  \begin{aligned}
  &\Pr[L > \ell_\beta + \eps_\beta]  + \Pr[ \ell_\beta >  L + \eps'] 
  \\&\le  
   8M \exp \left( -\frac{\Delta_\bx^2 \varepsilon'^2}{4} \min\{\beta, (\alpha - \beta)\} \tau M 
  \right) ~.
  \end{aligned}
  \end{align}
\end{lemma}




We leave the detailed proof of Lemma~\ref{lem:concen} in the full version.
To avoid linear regret, we desire to control $\Pr[L > \ell_\beta + \eps_\beta]  + \Pr[ \ell_\beta >  L + \eps']  \le O(1/T)$.
From Lemma~\ref{lem:concen}, we use the assumption on $\tau$ to bound the RHS of~\eqref{eq:L-beta-concen} by $8M\exp(-\ln(2K) \min\{\beta,(\alpha-\beta)\} M)$.
Then, it suffices to control 
\begin{align*}
  M \exp \lt( -\ln(2K) \min\{\beta,~\alpha-\beta\} M \rt) \le \fr{1}{T}.
\end{align*}
This takes the form of $M\exp(-Z^{-1}M) \le 1/T$ for some $Z$, which can be written as $M \ge Z\ln(M) + Z\ln(T)$.
One can show that a sufficient condition on $M$ to satisfy this inequality is
  $M \ge 2Z \ln\del{2ZT}$
where $Z = \fr{1}{\min\cbr{\beta,\alpha-\beta} \ln(2K)}$.

Define $\blue{\Delta_{\bmu}} := \max_{i\in[K]} \mu_* - \mu(i) \le 1$.
Assuming the condition on $M$ holds true, the following regret decomposition concludes the proof:
\begin{align}
  \!\!\!\! R^\pi_T (\vec{\mu}) &\le \EXP_\pi \! \bigg[ \sum_{i \in [K]}  (\mu_* - \mu(i)) n_T(i)  \mid  L + \varepsilon'+ \varepsilon_\beta 
  \ge  \hat{L}_\beta \ge L\bigg]  \nonumber \\
  & \quad + \Delta_{\bmu} T \left( \Pr[L > \hat{L}_\beta]  + \Pr[ \hat{L}_\beta >  L + \varepsilon' + \varepsilon_\beta] \right)
  \\
  &\le  (1+\lambda)  C(\vec{\mu}, L+\varepsilon'+ \varepsilon_\beta) \log T  + o(\log T)
  \nonumber
  \\&\quad + \Delta_{\bmu} T \cd 8 M \exp \left( -\tfrac{\Delta_\bx^2 \varepsilon'^2}{4} \min\{\beta, (\alpha - \beta)\} \tau M \right) 
  \nonumber
  \\
  &\le   (1+\lambda)  C(\vec{\mu}, L+\varepsilon'+ \varepsilon_\beta) \log T + o(\log T) \;, \nonumber
\end{align}
where 
we use 
Theorem~\ref{thm:upper}.

\section{Numerical Evaluation}
\label{sec:numerical}

\noindent{\bf Setup.}
For numerical experiments, 
we consider Lipschitz structure $\Phi(5)$ bandit setting with 6 arms,
embedding $\vec{x} = [0, 0.8, 0.85, 0.9, 0.95, 1]$ 
and time horizons $T = 10,000$. For each episode $m \in [M = 400]$, 
$\vec{\mu}_m \in \Phi (L)$ is independently generated by the following procedure: 
starting with $\mu_1 \in [0.05, 0.95]$ selected uniformly at random, for $i = 2, 3, ..., K$, select $\mu_i$ uniformly at random from $[\mu_{i-1}- L d(i-1,i) ,\mu_{i-1} + L d(i-1,i)] \cap  [0.05,0.95]$. 
Figures~\ref{fig:num:eg:L5} shows five $\vec{\mu}_m$'s generated from the generative procedure.
For every estimator (stated below), we use the same sequence of $\hat{L}_m$'s generated by $\pi(\infty)$ that uses no continuity structures.



\smallskip
\noindent{\bf Stable estimator $\hat{L}_\beta$.}
Figure~\ref{fig:num-re:estimate:L5} illustrates the comparison of accuracy between the four estimators on $L$: 
three $\hat{L}_\beta$'s with $(\beta, \varepsilon_\beta) \in \{ (0.5, 0.05)$, $(0.3, 0.05)$, $(0.1, 0.05) \}$ and $\max(\hat{L}_{m})$ which takes the maximum of $\hat{L}_m$'s estimated previously.
As expected, the most conservative estimation of $\max(\hat{L}_{m})$, which can theoretically explode up to $1/\Delta_{\vec{x}} =20$ in a finite number of episodes, has
monotonically increased estimation of $L$.
In contrast, each of $\hat{L}_\beta$'s stabilizes the estimation of Lipschitz constant $L$, once we have sufficient experiences, e.g., $M > 30$. Throughout the comparison of three $\hat{L}_\beta$'s, we can observe that the appropriate value of $\beta$ can bring benefits of Lipschitz continuity. Specifically the appropriate choice of $\beta = 0.3$ leads us to obtain an accurate and stable estimation of $L$, yielding the lower regret as shown in  Figure~\ref{fig:num-re:regret:L5}. 

\smallskip
\noindent{\bf Cumulative regret.}
Figure~\ref{fig:num-re:regret:L5} describes the cumulative regret under various estimators. We anticipate that the observations among 
the neighboring arms~$2$ to $6$ (i.e., $x \in [0.8,1]$), which are far from the embedding of arm $1$, are easily generalized to each other via Lipschitz continuity. 
As discussed in our theoretical analysis, 
we observe that the more accurate estimation of $L$ leads the greater reduction in regret. 
It suggests that there can be a significant difference according to $\beta$ we set,
which brings meaningful $\hat{L}_\beta$. In particular, the aggressive choice of $\beta=0.5$ gives a risky estimation $\hat{L}_\beta \lesssim L$; hence imposing a margin $\varepsilon_\beta$ may be helpful. Therefore, when we do not have information of structure, it is meaningful to transfer information of Lipschitz constant
to prior tasks by selecting the appropriate hyperparameter $\beta$ and $\varepsilon$.
We leave more detailed experimental descriptions in the full version.

\section{Conclusion} \label{sec:conclusion}

    We have investigated the role of \emph{transfer learning} of the unknown Lipschitz continuity. Our main contribution lies in our novel estimator $\hat L_\beta$, its information-theoretic optimality, and regret analysis when using $\hat L_\beta$ for future tasks, which is shown to be close to using the true unknown Lipschitz constant $L$ both theoretically and empirically.
    Our results serve as the first step towards transfer learning in structure bandits with tight instance-dependent regret guarantees, which opens up numerous fascinating research directions.
    For example, it would be interesting to extend  our setup to fully {\it adaptive} sequential transfer setting where the target accuracy $\eps$ is adjusted automatically as the learner solves more episodes. 
    Another direction is to assume that the embedding $\bx$ is not available where the learner must estimate it on the fly. 

\section{Acknowledgements}

This work was partly supported by Institute of Information \& communications Technology Planning \& Evaluation (IITP) grant funded by the Korea government (MSIT) (No. 2019-0-01906, Artificial Intelligence Graduate School Program (POSTECH)) and (No. 2021-0-00739, Development of Distributed/Cooperative AI based 5G+ Network Data Analytics Functions and Control Technology).


\clearpage

\bibliographystyle{IEEEtran}
\bibliography{ref}


\clearpage

\onecolumn

\appendices


\section{A Simplified version of Directed Exploration Learning Algorithm \cite{ok2018exploration}}

\begin{algorithm*}[!h]
    \caption{$\pi(L)$}
	\label{alg:corrected}
    \begin{algorithmic}
    \FOR{$t= 1,2,...$}
        \IF{$\exists i \in [K]$ s.t. $n_t(i)\leq\frac{\log{t}}{\log{\log{t}}}$\!\textsuperscript{\tred{$\dagger$}}} 
        \STATE \COMMENT{Estimation}
	         Play the most under-sampled arm $i_t \in \argmin_{i \in [K]} n_t(i)$ 
	    \ELSIF{
	    $\frac{\vec{\zeta}_t}{1+\lambda} \in \set{D}(\hat{\vec{\mu}}_t, L)$
	    } 
	    \STATE \COMMENT{Exploitation} Exploit the most under-sampled current best arm~$i_t \in 
        \argmin_{i \in \set{K}_*(\hat{\vec{\mu}}_t)}  n_t(i)$
        \ELSE  
        \STATE \COMMENT{Exploration}
           Compute $\vec{\eta}(\vec{\hat{\mu}}_t,  L)$ and set $\vec{\eta}_t$ as follows:
        \begin{align}
            \eta_t (i) = 
            \begin{cases}
            \min\{\log t \textsuperscript{\tred{$\dagger$}}, \vec{\eta}(i; \hat{\mu}_t, L)\},  & \text{if } i \in \set{K}_* (\vec{\mu}); \\
            (1+\lambda) \vec{\eta}(i; \hat{\mu}_t, L), & \text{otherwise.}
            \end{cases}
    	\end{align}
        \STATE Play the most under-explored arm $i_t 
        \in \argmax_{i \in [K]} \left( \eta_t(i) \log{t} - n_t(i) \right) $
        \ENDIF
	    \STATE Update statistics $\hat{\vec{\mu}}_{t+1}$ based on new reward $r_t$ corresponding to arm $i_t$:
		for each $i \in [K]$, set:
		\begin{align*}
    		n_{t+1} (i) = n_{t} (i) + \one[i_t = i]; 
    		\qquad \text{and} \qquad
    		\hat{\mu}_{t+1} (i) = \hat{\mu}_{t} (i) + \one[i_t = i]
    		 \left(\frac{r_t - \hat{\mu}_t(i)}{n_t(i) +1}\right).
     	\end{align*}
 		\ENDFOR
	   \medskip
	   \hrule
	\end{algorithmic} 
	{\tiny
\textsuperscript{\tred{$\dagger$}}
Positive constants 
can be multiplied to these terms
for stabilizing the empirical behavior
without harming the asymptotic optimality in Theorem~\ref{thm:upper}.
}
\end{algorithm*}

Let $\vec{\eta}(\vec{\mu}, L) \in \bar{\mathbb{R}}^{K}_+$ be a solution of LP \eqref{eq:opt}
where we set $\eta(i ; \vec{\mu}, L) = \infty$ for optimal arm $i \in \set{K}_* (\vec{\mu})$.
Then for the suboptimal arm~$i \notin \set{K}_* (\vec{\mu})$, 
$\vec{\eta}(i; \vec{\mu}, L) \log t$ provides a suggestion on minimal exploration at time $t$.
This motivates to an algorithm that keeps tracking the estimated LP solution $\vec{\eta} (\hat{\mu}_t, L)$ (if we knew $L$ a priori)
where $\hat{\mu}_t$ is the estimation of $\vec{\mu}$ at time~$t$.
Indeed, there have been a number of algorithms that use this framework to achieve the asymptotic lower bound in Theorem~\ref{thm:lower}
(if $L$ is known in advance); e.g., 
Optimal Sampling for Structured Bandit (OSSB)~\cite{Combes2017OSSB} 
and Directed Exploration Learning (DEL)~\cite{ok2018exploration}.
In this paper, for a given $L$,
we use Algorithm~\ref{alg:corrected}, denoted by $\pi(L)$, that is a simplified version of 
DEL algorithm
\cite{ok2018exploration}, which is originally designed 
for structured Markov decision process. 
We choose DEL algorithm as it has better empirical
behavior than OSSB thanks to the careful handling (such
as monotization) for cases where the assumptions for the
analysis are broken.

Algorithm~\ref{alg:corrected} consists of three phases: estimation, exploitation, and exploration.
The estimation phase ensures 
that every arm is sampled at least $\Omega(\log t / \log \log t)$.
This harms no regret asymptotically as $\log T / \log \log T = o(\log T)$, 
but it ensures concentrations of 
the estimates $\hat{\vec{\mu}}_t$
and $\vec{\eta}(\hat{\vec{\mu}}_t, L)$ to 
$\vec{\mu}$ and $\vec{\eta}({\vec{\mu}}, L)$, respectively. 
Let $\vec{\zeta}_t \in \mathbb{R}^K$ be the current rate of exploration at time $t$
such that ${\zeta}_t (i) \defeq \frac{n_t (i)}{\log t}$ for every $i \in [K]$. 
The algorithm exploits the current best arm
if the current exploration is statistically sufficient to identify the best arm, 
i.e., $\frac{\vec{\zeta}_t}{1+\lambda} \in \set{D}(\vec{\mu}, L)$
with some positive margin $\lambda > 0$.
If not, then it explores toward clipped LP solution ${\vec{\eta}}_t$ 
defined as:
\begin{align*}
\eta_t (i) \defeq
\begin{cases}
\min\{\log t, \vec{\eta}(i; \hat{\mu}_t, L)\},  & \text{if } i \in \set{K}_* (\vec{\mu}); \\
(1+\lambda) \vec{\eta}(i; \hat{\mu}_t, L), & \text{otherwise.}
\end{cases}
\;.
\end{align*}


\section{Proof of Lemma~\ref{lem:concen}}
\label{sec:proof-concen}
We begin with a concentration analysis on  $\hat{L}_m$ to $L_m$. 
Note that Assumption~\ref{asm:minimal} guarantees 
that every arm is played at least $\tau$ times in each episode. 
Hence, using Hoeffding's inequality, 
it follows that
for every $m \in [M]$, 
\begin{align} \label{eq:mu-concen}
 \Pr \left[ | \mu_m (i) - \hat{\mu}_m (i)  | \ge \varepsilon \right] \le 2 \exp\left( - 2\varepsilon^2  \tau \right)  \;,
\end{align}
which implies
\begin{align} \label{eq:Lm-concen}
  \begin{aligned}
&\Pr \left[ | L_m - \hat{L}_m  | \ge \varepsilon \right] 
\\&\le \Pr \left[\exists i \in [K]: |\hat\mu_m(i) - \mu_m(i)| \ge \frac{\varepsilon\Delta_\bx}{2} \right]
\\&\le 2 K \exp\left( - \frac{\Delta_\bx^2 \varepsilon^2  \tau}{2} \right)  ~.
\end{aligned}
\end{align}

For the following proofs, without loss of generality, we assume that $L_1 \ge \cdots \ge L_M$ and define $\xi_i = L - L_i$ be the Lipschitz gaps. 

\noindent{\bf Bound of $\Pr[L > \ell_\beta + \eps_\beta]$.}
Let us first show the upper confidence bound derivation. 
Let $[x]_+ = \max\cbr{0, x}$.
Recall that $\xi_m$ is nondecreasing by definition.
Let $S(i)$ be the $i$-th smallest element of the set $S$.
Let $m_\beta := \lceil\beta M \rceil$.
Define $\textbf{\cS} := \{S\subseteq[M]: |S| = M-m_\beta+1\}$.
Then,
\begin{align*}
  \PP\del{\kmax{m_\beta}_{m\in[M]}\hat L_{m} \le L - \eps_\beta}
  &\le \sum_{S\in \cS} \prod_{i=1}^{M-m_\beta+1} \PP(\hat L_{S(i)} \le L - \eps_\beta) 
  \\&= \sum_{S\in \cS} \prod_{i=1}^{M-m_\beta+1} \PP\del{\hat L_{S(i)} - L_{S(i)} \le - (\eps_\beta - \xi_{S(i)})}
  \\&\le \sum_{S\in \cS} \prod_{i=1}^{M-m_\beta+1} 1 \wedge 2K\exp(-\fr{\tau}{2}\Delta^2_\bx[\eps_\beta-\xi_{S(i)}]_+^2)~.
\end{align*}
Using $S(i) \le m_\beta+i-1\ \forall i \in [M-m_\beta+1]$, we have $\xi_{S(i)} \le \xi_{m_\beta + i - 1} $.
Then, the RHS above is bounded by 
\begin{align*}
  \sum_{S\in \cS} \prod_{i=1}^{M-m_\beta+1} 1 \wedge  2K\exp(-\fr{\tau}{2}\Delta_\bx^2[\eps_\beta-\xi_{m_\beta+i-1}]_+^2)
  &= \sum_{S\in \cS} \prod_{m=m_\beta}^{M} 1 \wedge  2K\exp(-\fr{\tau}{2} \Delta_\bx^2 [\eps_\beta-\xi_{m}]_+^2)
  \\&= \begin{pmatrix}
    M \\ M-m_\beta+1
  \end{pmatrix} \prod_{m=m_\beta}^{M} 1 \wedge  2K\exp(-\fr{\tau}{2}\Delta_\bx^2[\eps_\beta-\xi_{m}]_+^2).
\end{align*}
Note that
\begin{align*}
\prod_{m=m_\beta}^{M} 1\wedge 2K\exp(-\fr{\tau}{2}\Delta_\bx^2[\eps_\beta -\xi_{m}]_+^2)
  &\le \min_{s} \del{1 \wedge 2K\exp(-\fr{\tau}{2}\Delta_\bx^2  [\eps_\beta - s]_+^2)}^{ |\{m\ge m_\beta:~ \xi_m \le s\}| }
  \\&\sr{(a)}{\le} \del{ 1 \wedge 2K\exp(-\fr{\tau}{2}\Delta_\bx^2  (\eps_\beta - \eps_\alpha)^2)}^{\alpha M - m_\beta + 1}
  \\&\le \exp\lt( \lt( \ln(2K) -\fr{\tau}{2}\Delta_\bx^2  (\eps_\beta - \eps_\alpha)^2\rt)\cd (\alpha - \beta) M\rt) \;,
\end{align*}
where $(a)$ is by Assumption~\ref{asm:tight}.
Recalling the definition of $\varepsilon'$
and 
using the fact that ${{M}\choose{m_\beta}} \leq 2^M \leq \exp(M)$, we finally get:
\begin{align*}
  \PP\del{\kmax{m_\beta}_{m\in[M]}\hat L_{m} \le L - \eps_\beta}
&\le 
\exp\lt(
(1+
(\alpha-\beta)  \ln(2K)) M
-\fr{\tau}{2}\Delta_\bx^2  \eps'^2 (\alpha-\beta) M\rt)  \\
& \le
\exp\lt(
-\fr{\tau}{4}\Delta_\bx^2  \eps'^2 (\alpha-\beta) M\rt) \;,
\end{align*}
where the last inequality is from the assumption $\tau \ge \fr{4}{\Delta^2_\bx (\eps_\beta-\eps_\alpha)^2  }
\left(
 \ln(2K) + \frac{1}{\alpha - \beta}
\right)$.

\noindent{\bf Bound of $\Pr[\ell_\beta >  L + \eps']$.}
From the definition of $\ell_\beta$ in \eqref{eq:typical-L}, it follows that
\begin{align*}
  \PP\del{\kmax{m_\beta}_{m\in[M]}\hat L_{m} \ge L +\eps'}
  &\le \PP(\exists S \subseteq [M]: |S|=m_\beta\ \forall m\in S, \hat L_m - \eps' \ge L)
  \\&\le \begin{pmatrix}  M \\ {m_\beta}\end{pmatrix} \del{2K \exp\del{-\fr{\Delta_\bx^2 \eps'^2 \tau}{2}}}^{m_\beta}
  \\&\stackrel{(b)}{\le}  \exp\del{ \left(\frac{1}{\beta}+ \ln (2K) \right) \beta M -\fr{\tau}{2} \Delta_\bx^2 \eps'^2 \beta M }
  \\&\stackrel{(c)}{\le}  \exp\del{-\fr{\tau}{4} \Delta_\bx^2 \eps'^2 \beta M },
\end{align*}
where $(b)$ follows from 
the fact that ${{M}\choose{m_\beta}} \leq 2^M \leq \exp(M)$; and $(c)$ follows from our assumption on $\tau$.
\ep

\section{Proof of Theorem~\ref{thm:upper}}

We note that our analysis can be concluded with
any other algorithm than Algorithm~\ref{alg:corrected}
if it achieves the asymptotic optimality provided in 
Theorem~\ref{thm:upper}.
Algorithm~\ref{alg:corrected} is a simplification
of DEL algorithm in \cite{ok2018exploration}
originally designed for Markov decision process (MDP).
It is straightforward to correspond
the bandit problem with Lipschitz continuity 
to an MDP of single state and $K$ actions
with Lipchitz continuity.
Hence, the proof will be concluded
by Theorem~4 in \cite{ok2018exploration}
once we correspond the following linear programming
to the one in \eqref{eq:opt}:
\begin{subequations}
\label{eq:opt-unsim}
\begin{align}
\label{eq:objective-unsim}
\underset{
\vec{\eta} \succcurlyeq 0 
}{\TN{min}} & ~~  \sum_{i \notin \set{K}_* (\vec{\mu})}  \big(\mu_* - \mu(i) \big) \eta(i)   \\
  \TN{s.t.~} 
& \sum_{i \notin \set{K}_* (\vec{\mu})} \kl(\mu(i) \| \nu(i))  \eta(i) \ge 1 
~ \forall \vec{\nu} \in \Psi (\vec{\mu}, L) \;,
\label{eq:feasible-unsim}
\end{align}
\end{subequations}
where $\Psi (\vec{\mu}, L) \subset \Phi(L)$
is the set of confusing parameters to $\vec{\mu}$ defined
as 
\begin{align*}
\Psi(\vec{\mu}, L) :=
\left\{
\vec{\nu} \in \Phi(L) :
\set{K}_* (\vec{\mu}) \cap \set{K}_* (\vec{\nu}) = \emptyset ~~\text{and}~~ \mu(i)  = \nu(i) \; \forall i \in \set{K}_* (\vec{\mu})
\right\} \;.
\end{align*}
The correspondence is provided by Theorem~1 in \cite{Magureanu2014Lipschitz}.
This completes the proof.
\ep


\section{Proof of Theorem~\ref{thm:continuity}}

We will show that for fixed $\vec{\mu}$ and embedding $\vec{x}$, there exists $\delta$ satisfying~\eqref{eq:delta_cond} such that $|C(\vec{\mu}, L+\delta)-C(\vec{\mu}, L)| \rightarrow 0$ as $\delta \rightarrow 0$. Notice that $C(\vec{\mu}, L)$ is the minimal value for the optimization problem \eqref{eq:objective}\textendash\eqref{eq:feasible2} in Theorem~\ref{thm:lower}. The idea is to bound $C(\vec{\mu}, L+\delta)$ in terms of $C(\vec{\mu}, L)$. If we mistake the bandit model $\Phi(L)$ for $\Phi(L+\delta)$, we get $C(\vec{\mu}, L+\delta)$ by solving the optimization  \eqref{eq:objective}\textendash\eqref{eq:feasible2}, but with a different KL-divergence term in~\eqref{eq:feasible2}: $\kl(\mu(i) \| \nu^j(i; L+\delta))$. Here for ease of description, we abbreviate $\nu^{j}(i ; \vec{\mu}, L)$ to $\nu^j(i; L+\delta)$ since we consider a fixed $\vec{\mu}$.

By the definition of $\nu^j(i; L)$ mentioned in Theorem~\ref{thm:lower}, one can readily see that
$\kl(\mu(i) \| \nu^j(i; L+\delta))=\kl(\mu(i) \| \nu^j(i; L)) -\epsilon_{ij}$ for some $\epsilon_{ij}$, where $\epsilon_{ij} \rightarrow 0$ as $\delta \rightarrow 0$.
Now we claim that with $\delta$ satisfying~\eqref{eq:delta_cond}, $\epsilon_{ij} \leq 2\sqrt{DK}\delta\ \forall i, j \in [K]$. The rationale behind this claim is as follows:

\begin{align*}
& \kl(\mu(i)\| \nu^j(i; L+\delta)) \\
&= \mu(i) \log \left( \frac{\mu(i)}{\nu^j(i; L+\delta)}\right) + (1-\mu(i)) \log \left( \frac{1-\mu(i)}{1-\nu^j(i; L+\delta)}\right) \\
&= \mu(i) \log \left(\frac{\mu(i)}{\nu^j(i; L)}\cdot\frac{\nu^j(i; L)}{\nu^j(i; L+\delta)}\right) + (1-\mu(i)) \log \left( \frac{1-\mu(i)}{1-\nu^j(i; L)}\cdot\frac{1-\nu^j(i; L)}{1-\nu^j(i; L+\delta)}\right) \\
&= \kl(\mu(i)\| \nu^j(i; L)) + \kl(\nu^j(i; L)\| \nu^j(i; L+\delta)) + (\mu(i)-\nu^j(i; L))\log\left(\frac{\nu^j(i; L)}{1-\nu^j(i; L)}\cdot\frac{1-\nu^j(i; L+\delta)}{\nu^j(i; L+\delta)}\right) \\
&\stackrel{(a)}{\geq} \kl(\mu(i)\| \nu^j(i; L)) + (\mu(i)-\nu^j(i; L))\log\left(\frac{\nu^j(i; L)}{1-\nu^j(i; L)}\cdot\frac{1-\nu^j(i; L+\delta)}{\nu^j(i; L+\delta)}\right) \\
&\stackrel{(b)}{\geq} \kl(\mu(i)\| \nu^j(i; L)) + \min \! \Bigg\{  \! \log \! \left( \! \frac{\nu^j(i; L)}{1-\nu^j(i; L)}\cdot\frac{1-\nu^j(i; L)-\sqrt{DK}\delta}{\nu^j(i; L)+\sqrt{DK}\delta} \! \right), -\log \! \left( \! \frac{\nu^j(i; L)}{1-\nu^j(i; L)}\cdot\frac{1-\nu^j(i; L)+\sqrt{DK}\delta}{\nu^j(i; L)-\sqrt{DK}\delta} \! \right) \! \Bigg\},
\end{align*}
where $(a)$ follows from the non-negativity of KL-divergence; and $(b)$ holds by Proposition~\ref{prop:del-bound} that $\|\nu^j(i; L+\delta)-\nu^j(i; L)\| \leq \sqrt{DK}$ for sufficiently small $\delta$ and by dividing the second term into two subcases depending on the sign of $\mu(i)-\nu^j(i; L)$.
Now we claim that the second term of the last inequality is a function of $\delta$. To show this, let us focus on the first term inside the minimum operator. 
\begin{align*}
    &\log\left(\frac{\nu^j(i; L)}{1-\nu^j(i; L)}\cdot\frac{1-\nu^j(i; L)-\sqrt{DK}\delta}{\nu^j(i; L)+\sqrt{DK}\delta}\right) \\
    &=\left\{\log \nu^j(i; L) -\log \left(\nu^j(i; L)+\sqrt{DK}\delta\right)\right\} + \left\{\log\left(1-\nu^j(i; L)-\sqrt{DK}\delta\right)-\log\left(1-\nu^j(i; L)\right)\right\} \\
    &\geq \left\{-\frac{1}{\nu^j(i; L)}\sqrt{DK}\delta + \min_{x \in S }\frac{1}{x^2} \delta^2\right\} - \left\{\frac{1}{1-\nu^j(i; L)}\sqrt{DK}\delta -\min_{x \in S} \frac{1}{(1-x)^2}\delta^2\right\} \\
    &\geq -2\sqrt{DK}\delta \;.
\end{align*}
where we use Taylor series expansion and 
$S=[\nu^j(i; L), \nu^j(i; L)+\sqrt{DK}\delta]$.

Hence, for all $i, j \in [K]$, we obtain:
\begin{align}
    \kl(\mu(i)\| \nu^j(i; L+\delta)) \geq \kl(\mu(i)\| \nu^j(i; L)) - 2\sqrt{DK}\delta.
\end{align}
Let $C(\vec{\mu}, L)$ be the optimal solution for the bandit model $\Phi(L)$ and $\vec{\eta}_*(\vec{\mu}, L)\ (:=(\eta_*(1), \dots, 
\eta_*(K)))$ be its corresponding optimal allocation of samples. Then one can readily see that $\vec{\eta}_*(\vec{\mu}, L)\cdot\max_{i, j \in [K]} \frac{\kl(\mu(i)\| \nu^j(i; L))}{\kl(\mu(i)\| \nu^j(i; L)) - 2\sqrt{DK}\delta}$ is \emph{feasible} for the optimization problem \eqref{eq:objective}\textendash\eqref{eq:feasible2} for the bandit model $\Phi(L+\delta)$. As a consequence of this result,
\begin{align*}
    & C(\vec{\mu}, L+\delta) \leq C(\vec{\mu}, L)\cdot \max_{i, j \in [K]} \frac{\kl(\mu(i)\| \nu^j(i; L))}{\kl(\mu(i)\| \nu^j(i; L)) - 2\sqrt{DK}\delta}.
\end{align*}
Also, as mentioned in Section~\ref{sec:L-impact}, we note that $C(\vec{\mu}, L+\delta) \geq C(\vec{\mu}, L)$ due to the size of the feasible set. Hence we see that $C(\vec{\mu}, L+\delta) \rightarrow C(\vec{\mu}, L)$
as $\delta \rightarrow 0$. This completes the proof of continuity.

\begin{proposition} \label{prop:del-bound}
If $\delta$ satisfies~\eqref{eq:delta_cond},
we have $\|\nu^j(i; L+\delta)-\nu^j(i; L)\| \leq \sqrt{DK}$.
\end{proposition}
\begin{proof}
$\vec{\nu}^{j}$ in~\eqref{eq:feasible2} is the minimal perturbation 
to make (originally suboptimal) arm~$j \notin \set{K}_*(\vec{\mu})$ best in $\vec{\nu}^j$,
while preserving the Lipschitz continuity, i.e., $\vec{\nu}^{j} \in \Phi(L)$.
Specifically, we construct a confusing parameter for $\Phi(L)$ as follows. First, fix $j \in [K]$ such that $j \neq i^*$. Define $j$th confusing parameter as $\vec{\nu}^j(L):=(\nu^j(1; L), \dots, \nu^j(K; L))$, where 
$\nu^j(i; L) = \max\{\mu(i), \mu_{*}-L\|x(i) - x(j)\|\}$ such that $\nu^j(j; L) = \mu_{*}$.
Note that for confusing parameter $\vec{\nu}^j(L)$, we get $\nu^j(i; L) = \mu(i)$ if $\mu(i) \geq \mu_{*}-L\|x(i) - x(j)\|$, i.e., $\|x(i) - x(j)\| \geq \frac{1}{L}(\mu_{*}-\mu(i))$.

Now consider another confusing parameter for the perturbed bandit model $\Phi(L+\delta)$, say $\vec{\nu}^j(L+\delta)$ where $\nu^{j}(i; L+\delta) = \max\{\mu(i), \mu_{*}-(L+\delta)\|x(i) - x(j)\|\}$ such that $\nu^{j}(j; L+\delta) = \mu_{*}$. Also, note that if $\|x(i) - x(j)\| \geq \frac{1}{L+\delta}(\mu_{*}-\mu(i))$, we obtain $\nu^j(i; L+\delta) = \mu(i)$.From this, we observe that if there exists $i$ such that $\|x(i) - x(j)\| < \frac{1}{L}(\mu_{*}-\mu(i))$ and $\|x(i) - x(j)\| \geq \frac{1}{L+\delta}(\mu_{*}-\mu(i))$, then $\nu^j(i; L) = \mu_{*}-L\|x(i) - x(j)\|$, while $\nu^j(i; L+\delta) = \mu(i)$. Hence for fixed $(i, j)$, the perturbation of the bandit model by $\delta$ incurs a different behavior of the two confusing parameters. In order not for this to happen, $\delta$ has to be sufficiently small. Specifically, here is how we can set $\delta$.

Let $\zeta := (\mu_{*}-\mu(i))-L\|x(i) - x(j)\|$. Then one can express $(L+\delta)\|x(i) - x(j)\|$ as $(\mu_{*}-\mu(i))-\zeta+\delta\|x(i) - x(j)\|$. As long as we set $\delta < \frac{\zeta}{\|x(i) - x(j)\|}\ (=\frac{\mu_{*}-\mu(i)}{\|x(i) - x(j)\|}-L)$, $(L+\delta)\|x(i) - x(j)\| < (\mu_{*}-\mu(i))$. Hence, for any fixed $i, j \in [K]$, if we set
\begin{align*}
    \delta < \min_{i, j \in [K]}\left\{\frac{\mu_{*}-\mu(i)}{\|x(i) - x(j)\|}-L\right\},
\end{align*}
the two confusing parameters for the bandit models $\Phi(L)$ and $\Phi(L+\delta)$ exhibit exactly the same behavior. Let us now quantify how much deviation of each arm's mean reward occurs due to the perturbation of the bandit model by $\delta$. Since we set $\delta$ as above, for both $\Phi(L)$ and $\Phi(L+\delta)$, arm $i$'s mean reward is either $\mu(i)$ or $\mu_{*}-L\|x(i) - x(j)\|$ (or $\mu_{*}-(L+\delta)\|x(i) - x(j)\|$ respectively) for all $K-1$ confusing parameters. With this, one can see that the difference in each arm's mean reward is then up to $\delta\|x(i) - x(j)\| <\sqrt{DK}\delta$.
\end{proof}

\section{Proof of Theorem~\ref{thm:sample-lower}}
\label{sec:pf-sample-lower}

Let $n_{m}(i)$ denote the number of playing arm $i \in [K]$ in episode $m \in [M]$ with slight abuse of notation.
Let $\set{F}$ be the sigma-field
of observations in $M$ episodes
of length $T$, in which 
Assumption~\ref{asm:minimal} holds, i.e., $n_m (i) \ge \tau$ for every episode $m \in [M]$ and arm $i \in [K]$.
Then, the uniformly good estimator $\hat{L}$
verifies \eqref{eq:concen-L-required}
for $(\vec{\mu}_m)_{m \in [M]}$ satisfying Assumption~\ref{asm:tight}.
Let $\Pr$ and $\Pr'$ be the probability measures on $\set{F}$
w.r.t. $\set{M}$
and $\set{M}'$, respectively.
Similarly, denote the expectations
 on $\set{F}$
w.r.t. $\set{M}$
and $\set{M}'$ by $\EXP$ and $\EXP'$ respectively.
We use a change-of-measure argument
which compares two sequences of $M$ parameters, denoted by 
$\set{M} := (\vec{\mu}_m)_{m \in [M]}$ s.t.  $\vec{\mu}_m \in
\Phi (L)$
and $\set{M}' := (\vec{\nu}_m)_{m \in [M]}$
s.t.  $\vec{\nu}_m \in \Phi (L')$.
We will construct 
$\set{M}$ and $\set{M}'$
to conclude the proof using the following lemma:
\begin{lemma}[Lemma~19 in \cite{kaufmann2016complexity}]
\label{lem:change}
For every event $\set{E} \in \set{F}$,
\begin{align}
    \EXP [\set{G}]  \ge \kl(\Pr [ \set{E}] \| \Pr'[ \set{E}]) \;,
\end{align}
where $\set{G}$ is the log-likelihood ratio of $\set{M}$
to $\set{M}'$ defined as:
\begin{align*}
\set{G} := 
 \sum_{m \in [M]}
\sum_{i \in [K]}  n_{m}(i) \log \frac{\mu_m (i)}{\nu_m (i)} \;.
\end{align*}
\end{lemma}

Let $i'$ be an arm such that $\min_{i \neq i'} d(i, i') = \Delta_{\vec{x}}$. 
For some $c \in (0,1)$, we consider $\set{M}$ such that for each $m \in [M]$,
\begin{align*}
\mu_m(i)= 
\begin{cases} 
c  + L \Delta_{\vec{x}}  & \quad \text{if $i = i'$}  \\
c & \quad \text{otherwise}
\end{cases}    \;. 
\end{align*}
Note that $L_m:= \max_{i  \neq j \in [K]} \frac{|\mu_m(i) - \mu_m (j)|}{d(i,j)}  =  L$ for all $m \in [M]$, i.e., $\vec{\mu}_m \in \Phi(L)$. In addition, $\set{M}$ verifies
Assumption~\ref{asm:tight} for $L$, $\eps_\alpha$ and $\alpha$.
We now construct a perturbation $\set{M}'$
which verifies Assumption~\ref{asm:tight} for $L'= L+\eps$, $\eps_\alpha$ and $\alpha$.
For each $m  \in [ \lceil \alpha M \rceil]$, 
\begin{align} \label{eq:const-M'}
    \nu_m(i) = 
    \begin{cases}
    \mu_m(i') + (\eps - \eps_\alpha) \Delta_{\vec{x}}  & \quad \text{if $i = i'$}  \\
    \mu_m(i) & \quad \text{otherwise}
    \end{cases} \;,
\end{align}
which implies $L'_m := \max_{i  \neq j} \frac{|\nu(i) - \nu(j)|}{d(i,j)} = L+ (\eps - \eps_\alpha)$ 
due to the construction of $\vec{\mu}_m$:
for $i \neq i'$, 
\begin{align*}
\nu_m (i') - \nu_m(i)
&= (\mu_m(i') + (\eps - \eps_\alpha) \Delta_{\vec{x}}) - \mu_m(i)  \\
&=  \mu_m(i)+ L \Delta_{\vec{x}}
+ \eps \Delta_{\vec{x}} - \mu_m(i)  = (L+ \eps - \eps_\alpha) \Delta_{\vec{x}} \;.
\end{align*}
For the rest, i.e., $m \in [M] \setminus [\lceil \alpha M  \rceil]$, we set $\vec{\nu}_m = \vec{\mu}_m$.
Hence, $\set{M}'$ verifies Assumption~\ref{asm:tight}
for $L' := L + \eps$, $\alpha$ and $\eps_\alpha$.
Assume that
for each $m \in [M]$, $n_m(i') = \tau$ 
and $n_m(i) \ge \tau$ if $i \neq i'$.
This implies Assumption~\ref{asm:minimal}.
Note that
$\EXP[\set{G}] = \sum_{m \in [M]}
\sum_{i \in [K]} \EXP[n_m (i)] \kl (\mu_m(i) \| \nu_m(i))$
thanks to Markov property of bandit.
With the construction of  $\set{M}$ and $\set{M}'$,
it follows that 
\begin{align*}
    \EXP [\set{G}]
  &  \le  
    \tau 
    \left( 
    \sum_{m = 1}^{\lceil \alpha M \rceil}
    \kl( \mu_m (i') \| \nu_m (i'))
 \right) \\
 &\le 
 \tau 
 \left( 
 \sum_{m = 1}^{\lceil \alpha M \rceil}
\frac{(\mu_m(i') - \nu_m(i'))^2}{\nu_m(i') (1-\nu_m(i'))}    
 \right) \;,
\end{align*}
where for the last inequality, we use the fact that $\KL(\mu \| \nu) \le  \set{X}^2(\mu, \nu) =  \frac{(\mu - \nu)^2}{\nu (1-\nu)}$,
c.f,. Lemma~2.7 in \cite{tsybakov2008introduction}.
It is not hard to select constant $c \in (0,1)$
such that 
\begin{align*}
     \tau 
 \left( 
 \sum_{m = 1}^{\lceil \alpha M \rceil}
\frac{(\mu_m(i') - \nu_m(i'))^2}{\nu_m(i')  (1- \nu_m (i'))}
 \right)
 &= \Omega \left(
 \tau 
 \left( 
 \sum_{m = 1}^{\lceil \alpha M \rceil}
{(\mu_m(i') - \nu_m(i'))^2} 
 \right)
 \right) \;.
\end{align*}
With such choice of $c$
and the construction of $\vec{\nu}_m$ in \eqref{eq:const-M'}, we obtain
\begin{align}
    \EXP [\set{G}]
= \Omega \left(
 \tau 
  \alpha M  \Delta_{\vec{x}}^2(\eps -  \eps_\alpha)^2
 \right)
 \;. \label{eq:change-measure-1}
\end{align}

To complete the proof using Lemma~\ref{lem:change},
we define an event $\set{E} = \{\hat{L}  \in [L, L+\eps]\}$,
and its complement $\set{E}'$.
Under Assumption~\ref{asm:minimal} with $\tau > 0$
and the supposition that estimator $\hat{L}$ is uniformly good
for $(\alpha, \eps_\alpha, \tau, \eps)$,
we have $\Pr [\set{E}'] = o\left(\frac{\log T}{T} \right)$
and further 
\begin{align}
\Pr'[\set{E}] & = \Pr'[\hat{L}  \in [L, L+\eps]] \nonumber\\
&= \Pr'[\hat{L} \in [L'- \eps, L']]
= o \left(
\frac{\log T}{T} 
\right)\;, \nonumber 
\end{align}
where the last equality is from the construction of $\set{M}'$
verifying  Assumption~\ref{asm:tight} for $L' = L+\eps$, $\alpha$, and $\eps_\alpha$, i.e., 
\begin{align*}
    \Pr'[\hat{L} \in [L'- \eps, L']]
    \le 
     \Pr'[\hat{L} < L']
     + 
    \Pr'[\hat{L} >L'+\eps]  = o\left(\frac{\log T}{T} \right) \;.
\end{align*}
From this, it follows that
\begin{align}
\kl (\Pr [\set{E}] \|\Pr' [\set{E}] ) 
=  - \log \left( o \left( \frac{\log T}{T} \right) \right)
= O(\log T) \;.
    \label{eq:change-measure-2}
\end{align}
Therefore, combining \eqref{eq:change-measure-1}
and \eqref{eq:change-measure-2},
Lemma~\ref{lem:change} concludes the proof of Theorem~\ref{thm:sample-lower}. 
\ep

\section{Additional Experiments}

\subsection{Numerical evidence of risk simulation}


\begin{figure*}
    \centering
    \subfigure[Histogram of $\hat{L}_{T = 50k}$]{
        \includegraphics[width=0.31\textwidth]{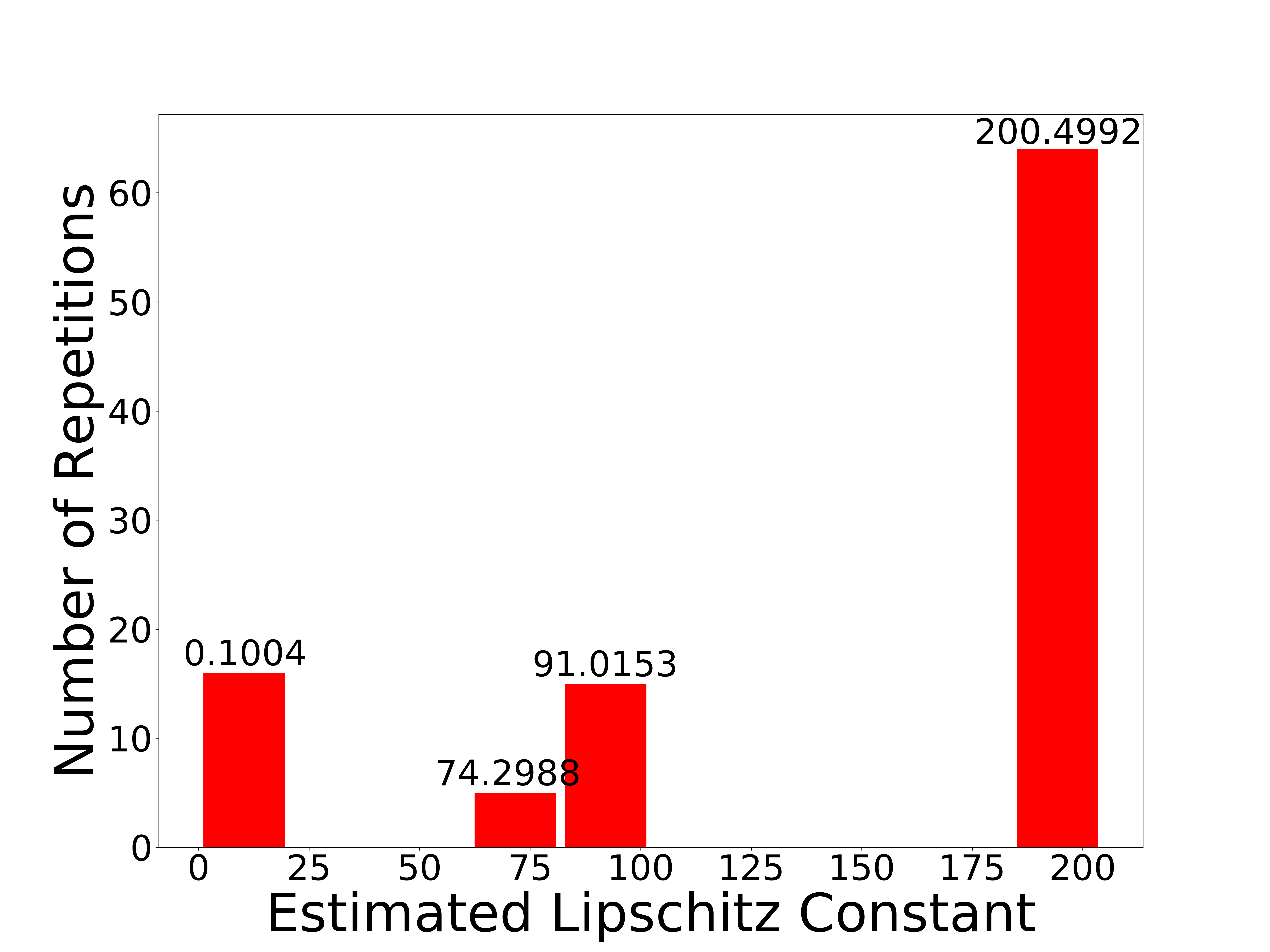}
        \label{fig:risk:hist}
        }
    \subfigure[Arm pulls $n_T(i)$'s at $T = 50k$]{
        \includegraphics[width=0.31\textwidth]{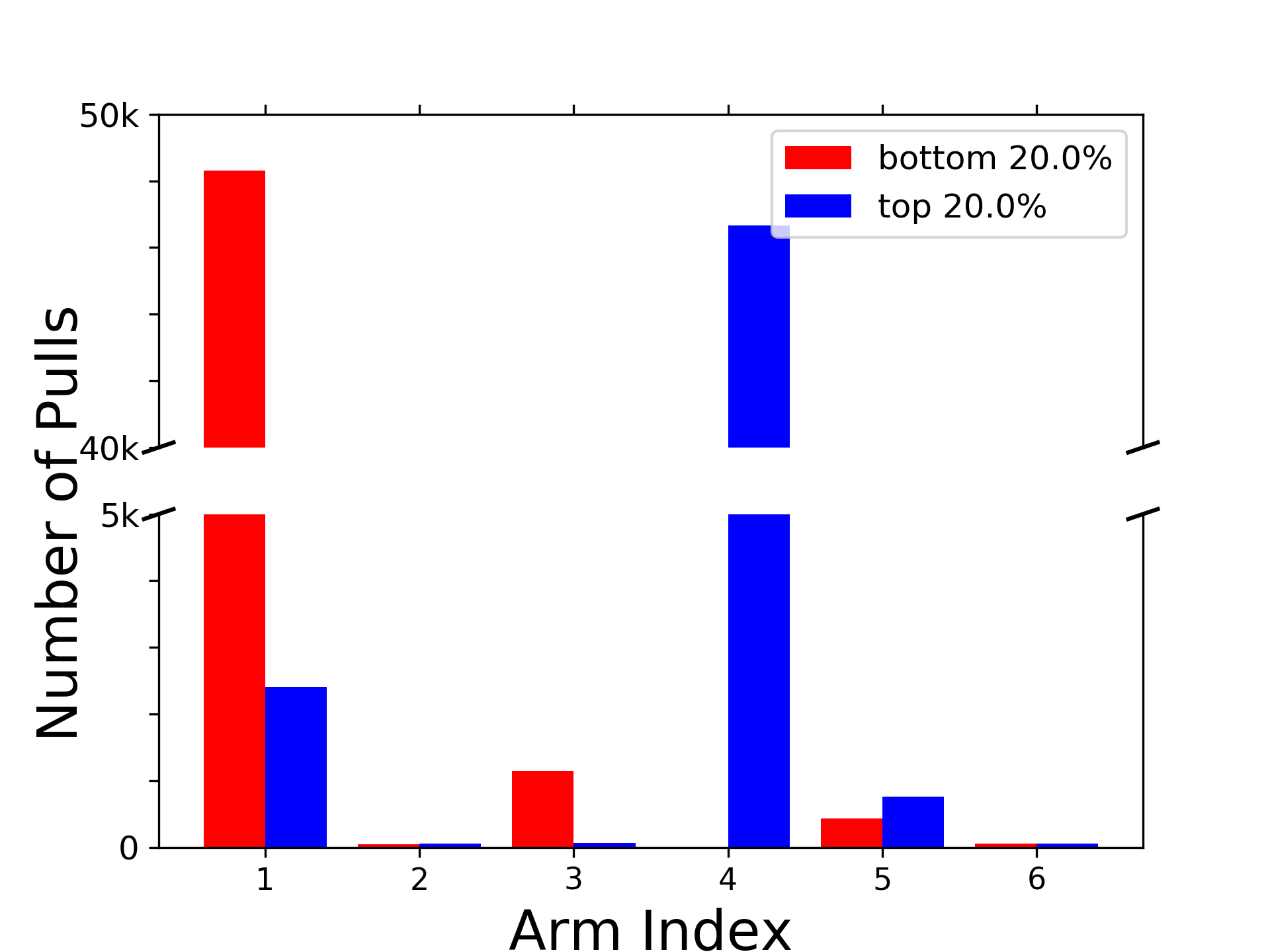}
        \label{fig:risk:pull}
        }
    \subfigure[Estimated $\hat{L}_t$ over $t$]{
        \includegraphics[width=0.31\textwidth]{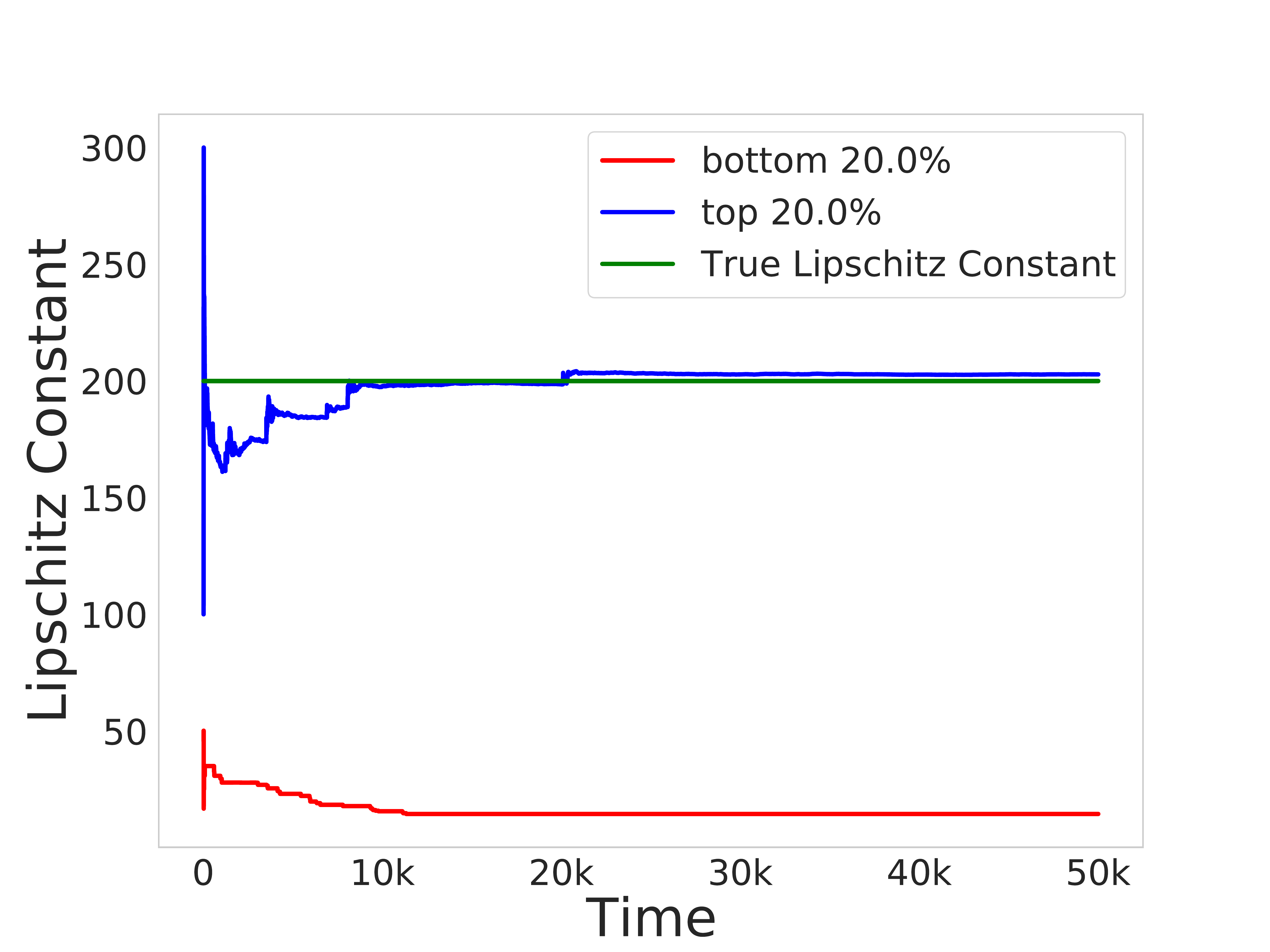}
        \label{fig:risk:est}
        }
    \caption{Behavior of $\pi(\hat{L}_{t})$
     for given $\vec{\mu}$ and $\vec{x}$ shown in Figure~\ref{fig:risk:para}.} 
\end{figure*}

We provide additional experiments with respect to a numerical evidence on a risk  
presented in Section~\ref{sec:L-impact}.
It is worth closely investigating the failure of $\pi(\hat{L}_t)$ with Figure~\ref{fig:risk}.
Figure~\ref{fig:risk:hist} presents the histogram of $\hat{L}_T$ at $T = 50k$.
A sustainable portion of sample paths
($16$ out of $100$; the first bin in Figure~\ref{fig:risk:hist})
has the estimation $\hat{L}_T$ concentrated around the second steepest slope ($0.1$) in Figure~\ref{fig:risk:para}.
Figure~\ref{fig:risk:pull}~and~\ref{fig:risk:est}
compare two groups of sample paths
with the bottom-$20$\% or top-$20$\% values of $\hat{L}_{T}$ at $T = 50k$ in Figure~\ref{fig:risk:hist}.
As shown in Figure~\ref{fig:risk:pull},
the bottom-$20$\% group with under-estimated $\hat{L}_t$
misidentifies the second best arm~$1$ as the best arm
and mainly contributes the linear regret.
An explanation on 
such a catastrophic failures
is provided in Figure~\ref{fig:risk:est} where 
the bottom-$20$\% group can hardly recover
from the estimation error in $\hat{L}_t$.
Suppose that at certain iteration $t$, the second best arm~$1$ (accidentally) has a much higher empirical mean than the others, i.e., $\hat{\mu}(1) \gg \max_{i = 2,..., 6}\hat{\mu}(i) \approx 0$.
Then, the value of $\hat{L}_t$ is much lower than $L$ as arm~$1$ is far from every other point.
The underestimated $\hat{L}_t$ forces the algorithm overgeneralize and excessively reduces the exploration rate on arms~$2$-$6$ and also the chance of correcting $\hat{L}_t$.
The failure of $\pi(\hat{L}_t)$ 
simultaneously learning structure and minimizing regret 
motivates us to study a scenario of transfer learning.

\subsection{Numerical evaluation}
\begin{figure*}[!ht]
    \centering
    \subfigure[Incidences of $\vec{\mu}_m$'s ($L = 0.5$)]{
        \includegraphics[width=0.31\textwidth]{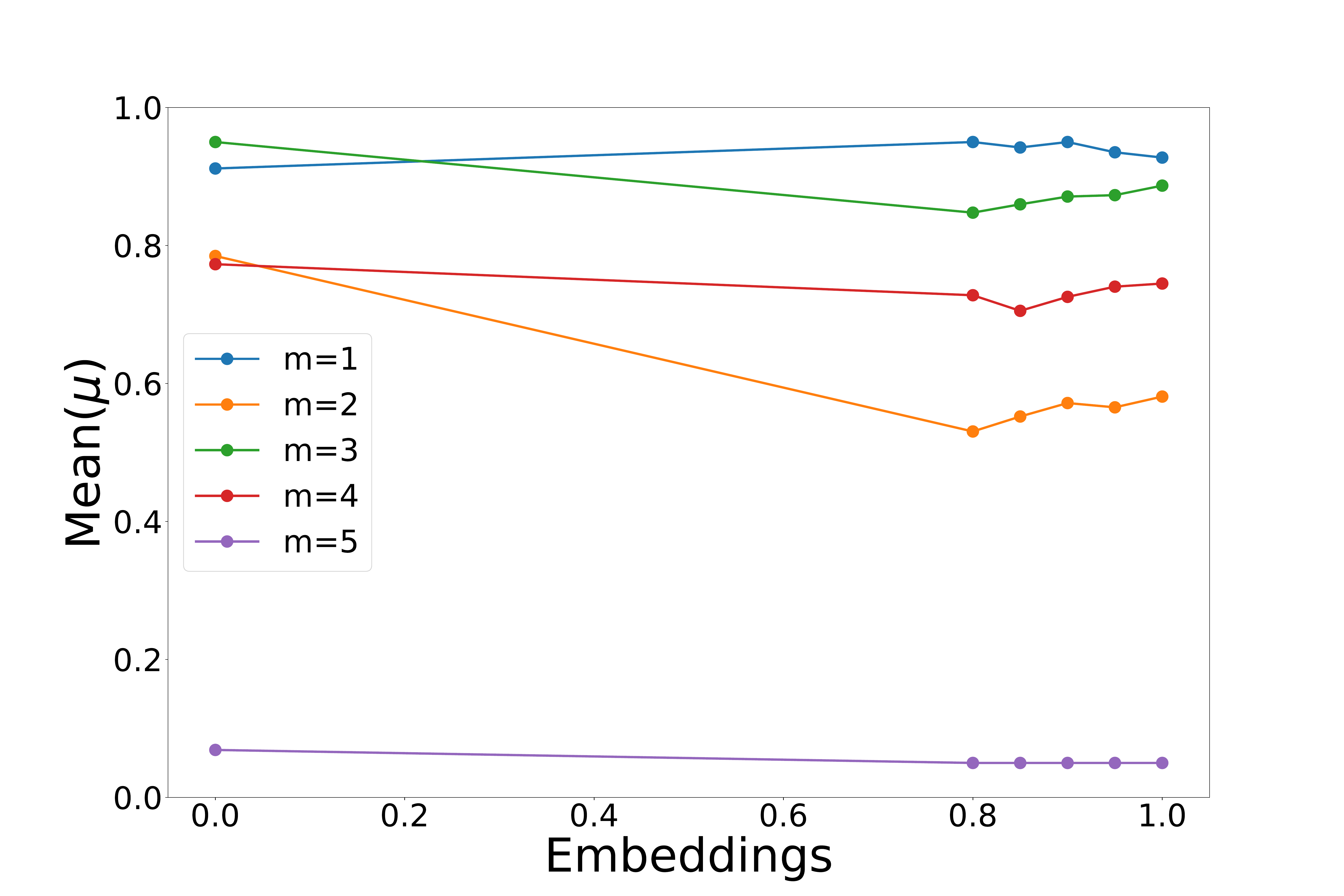}
        \label{fig:num:eg:L0.5}
        }
    \subfigure[Histogram of $L_m$'s ($L = 0.5$)]{
        \includegraphics[width=0.31\textwidth]{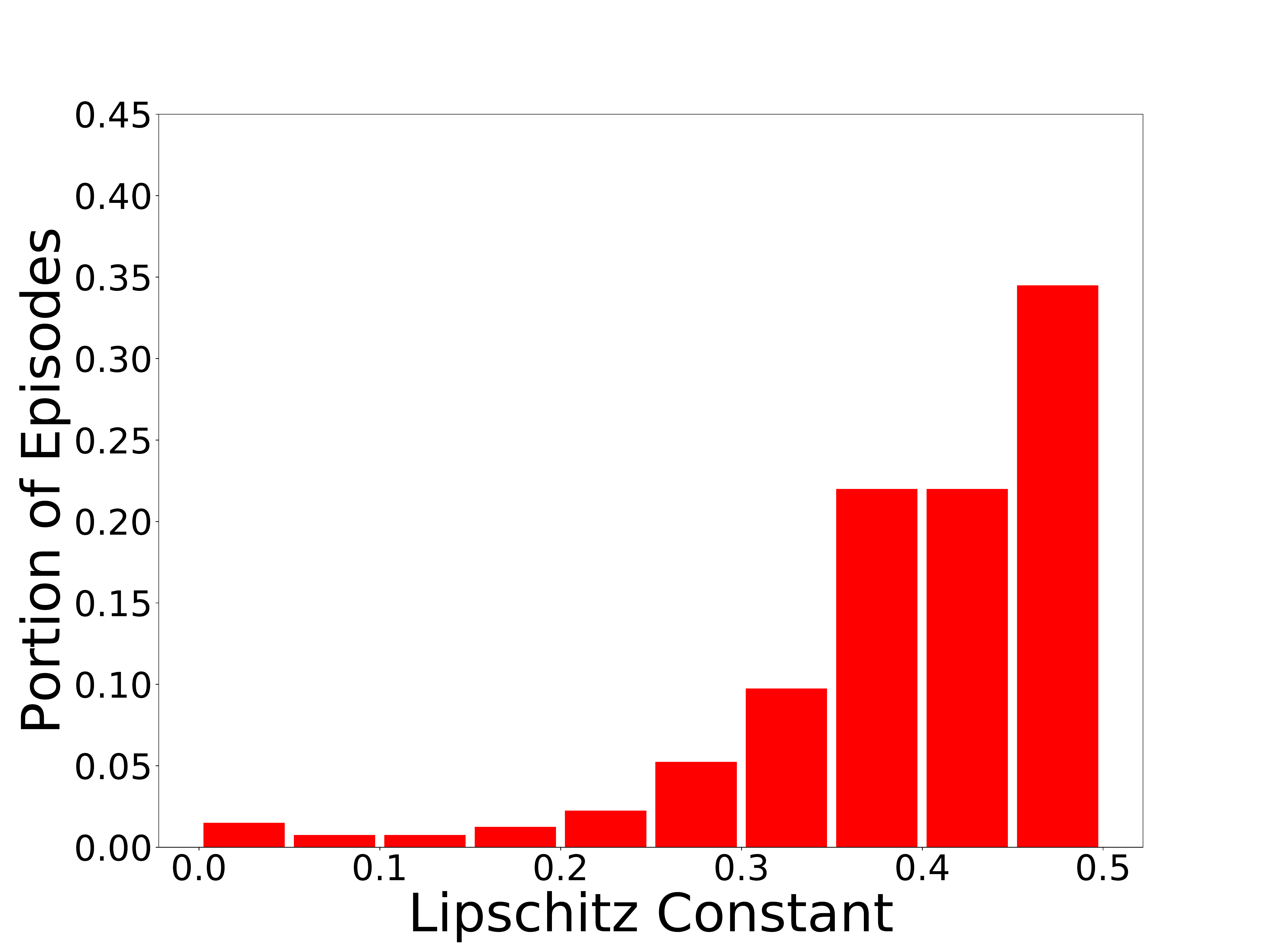}
        \label{fig:num:hist:L0.5}
        }
    \subfigure[Histogram of $\hat{L}_m$'s ($L = 0.5$)]{
        \includegraphics[width=0.31\textwidth]{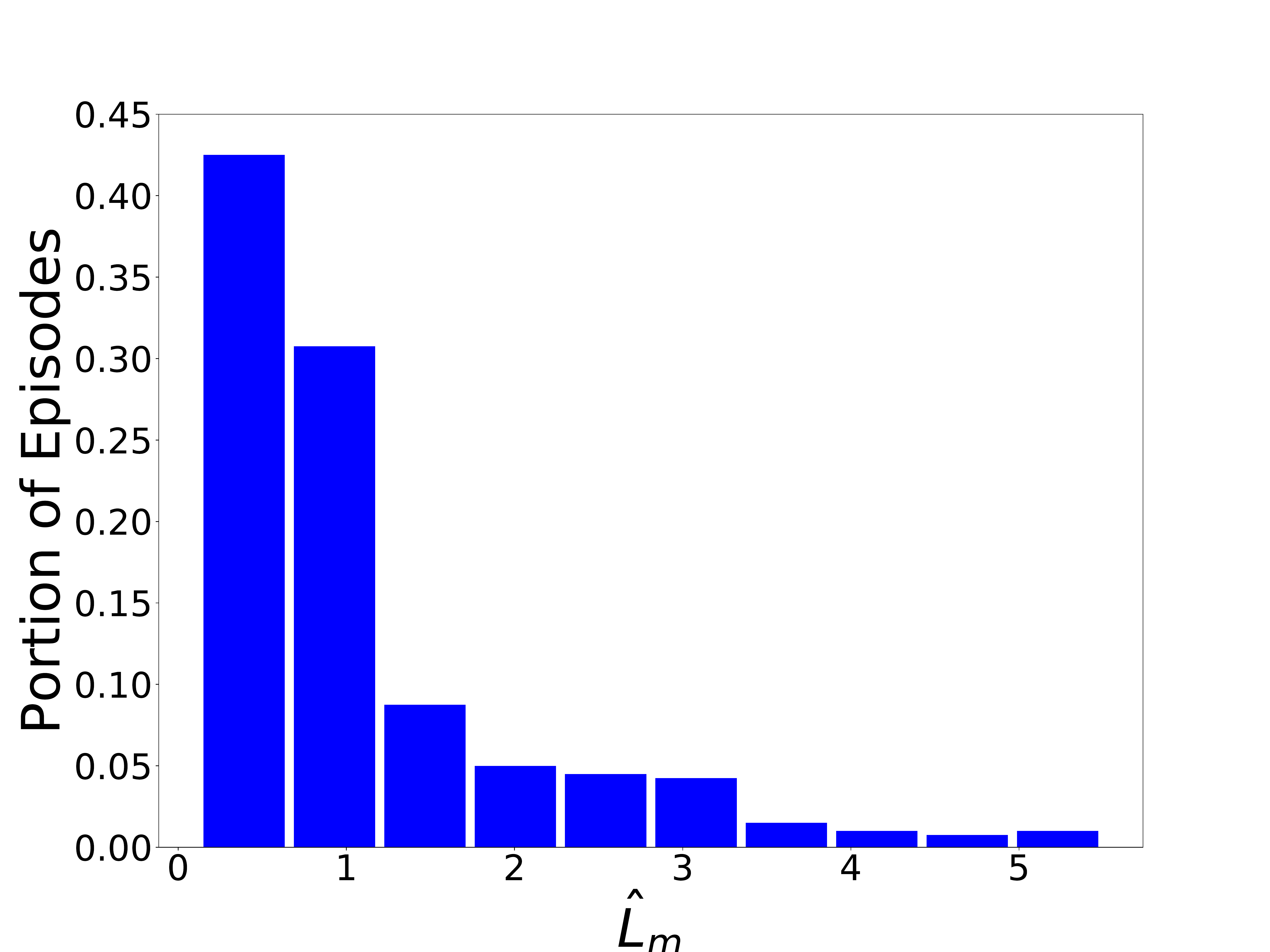}
        \label{fig:num:em-hist:L0.5}
        }
    \\
    \subfigure[Incidences of $\vec{\mu}_m$'s ($L = 5$)]{
        \includegraphics[width=0.31\textwidth]{simul_gene5.pdf}
        \label{fig:num:eg:L5-2}
        }
    \subfigure[Histogram of $L_m$'s ($L = 5$)]{
        \includegraphics[width=0.31\textwidth]{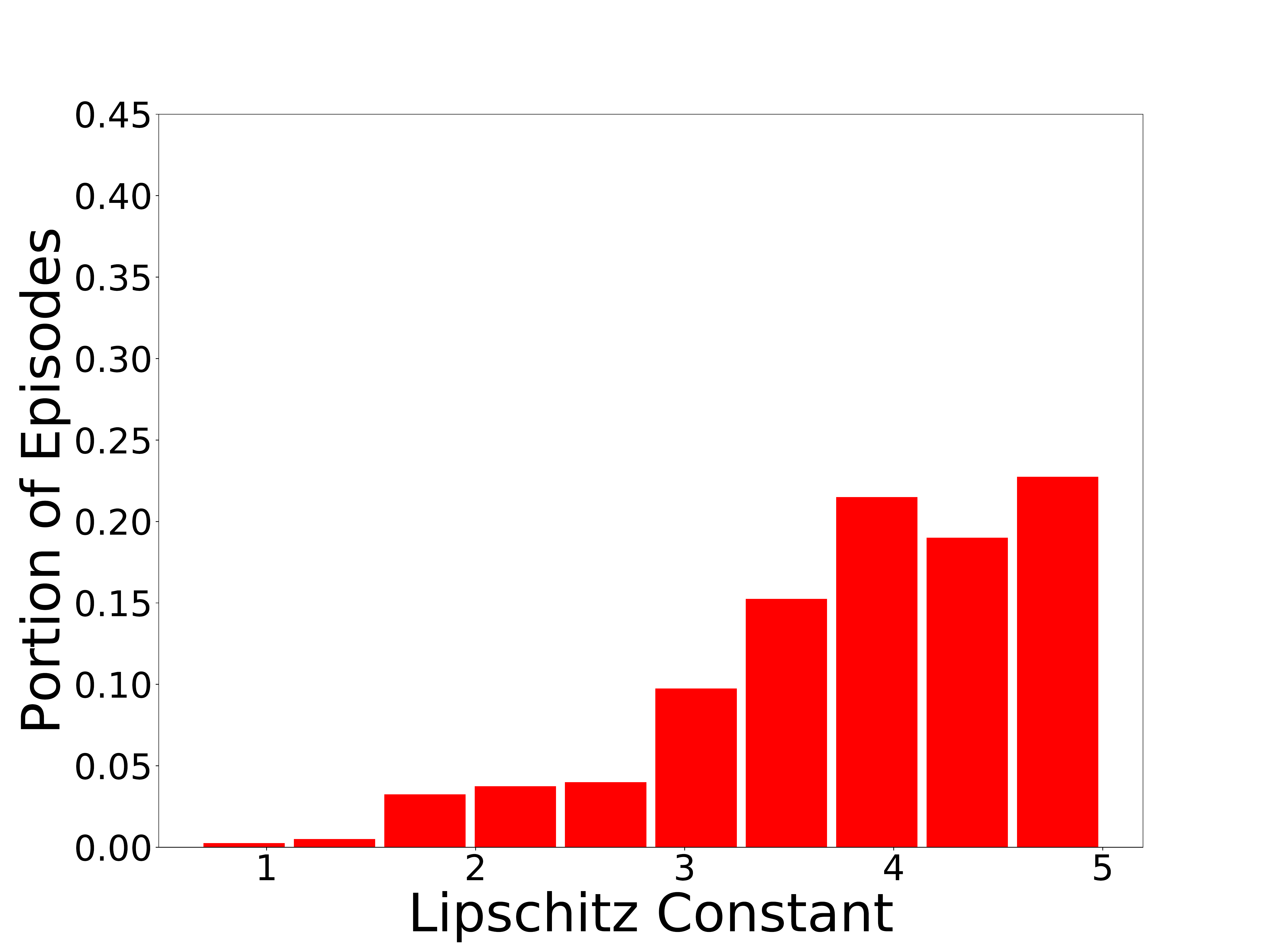}
        \label{fig:num:hist:L5}
        }
    \subfigure[Histogram of $\hat{L}_m$'s ($L = 5$)]{
        \includegraphics[width=0.31\textwidth]{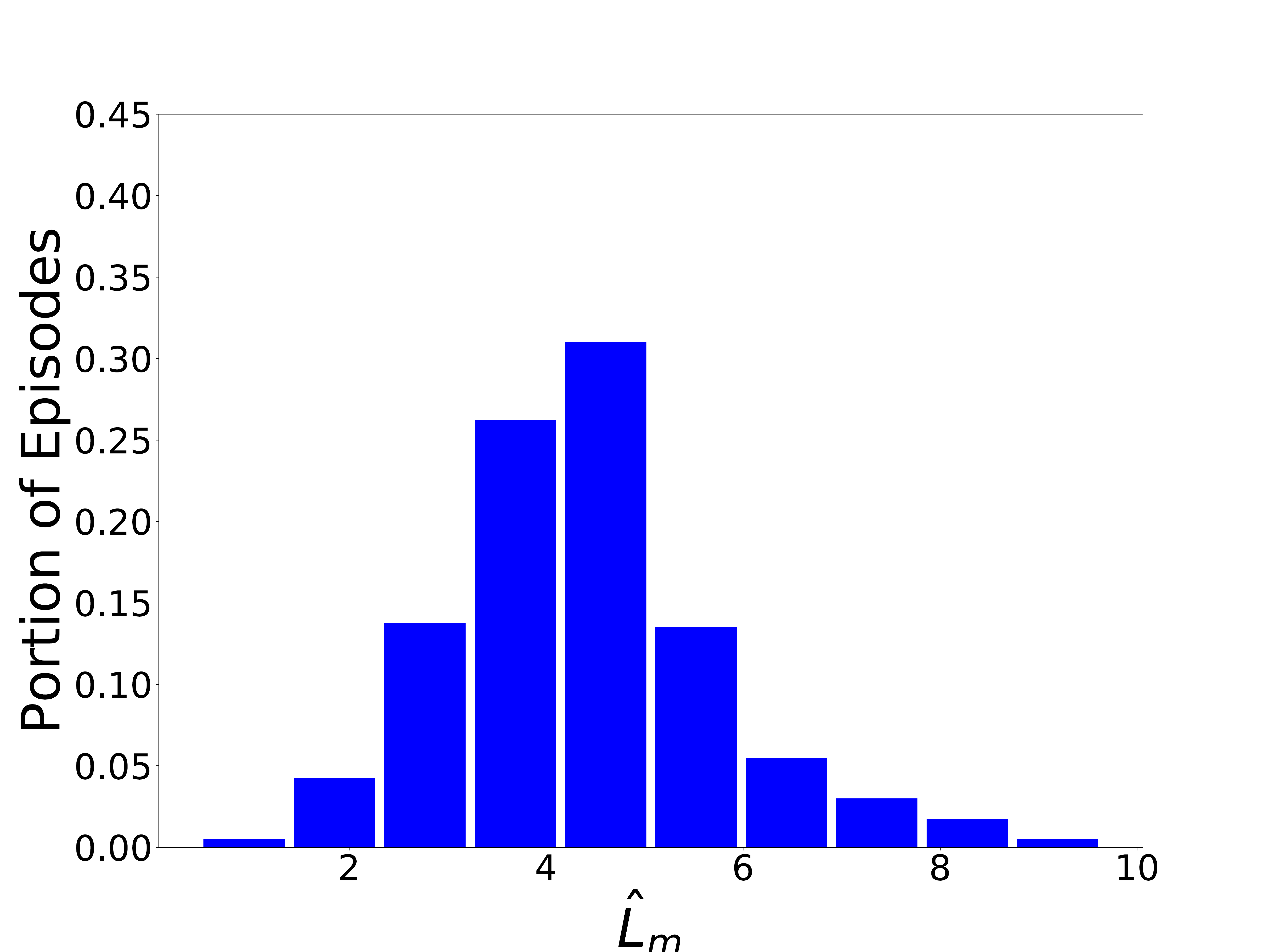}
         \label{fig:num:em-hist:L5}
         }
    \caption{Examples and statistics
    of experienced tasks generated by the procedure in Section~\ref{sec:numerical} with $L \in\{ 0.5, 5  \}$} 
    \label{fig:num:L}
\end{figure*}

\begin{figure*}[!h]
    \centering
    \subfigure[Evolution of estimation over episodes
     ($L = 0.5$)]{
        \includegraphics[width=0.42\textwidth]{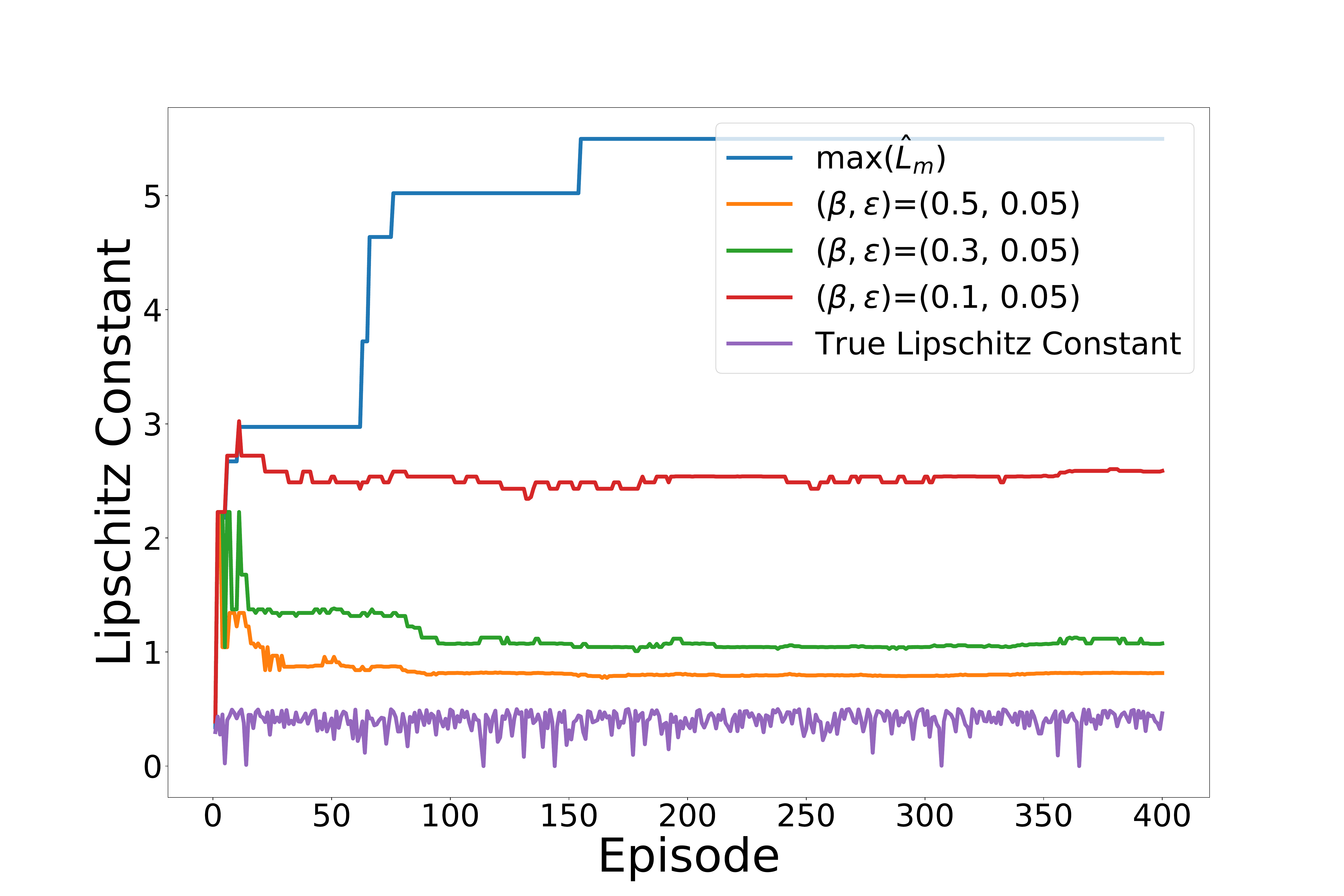}
        \label{fig:num-re:estimate:L0.5}
        }
    \subfigure[Cumulative regret over episodes
     ($L = 0.5$)]{
        \includegraphics[width=0.42\textwidth]{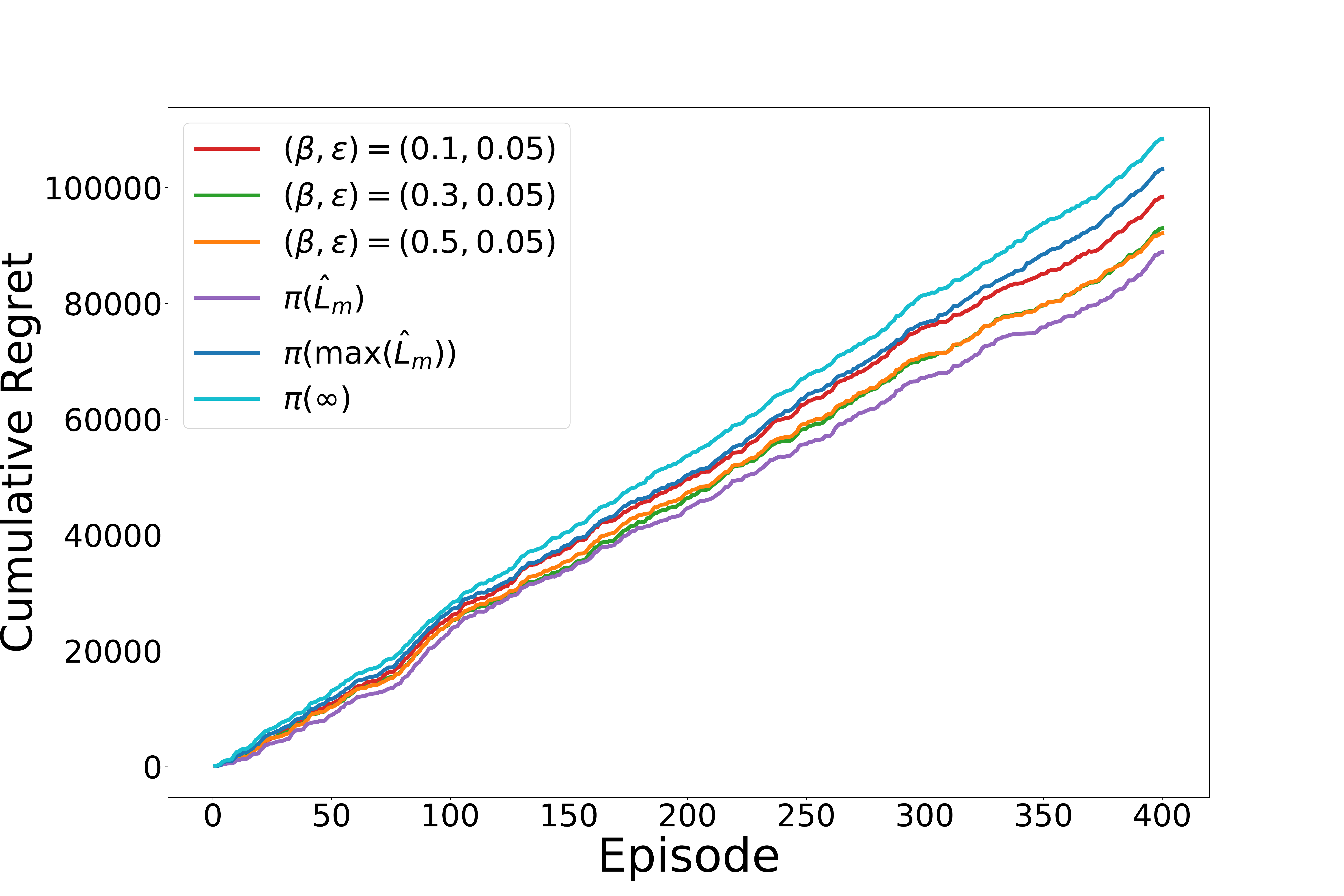}
        \label{fig:num-re:regret:L0.5}
        }
    \\
    \subfigure[Evolution of estimation over episodes
     ($L = 5$)]{
        \includegraphics[width=0.42\textwidth]{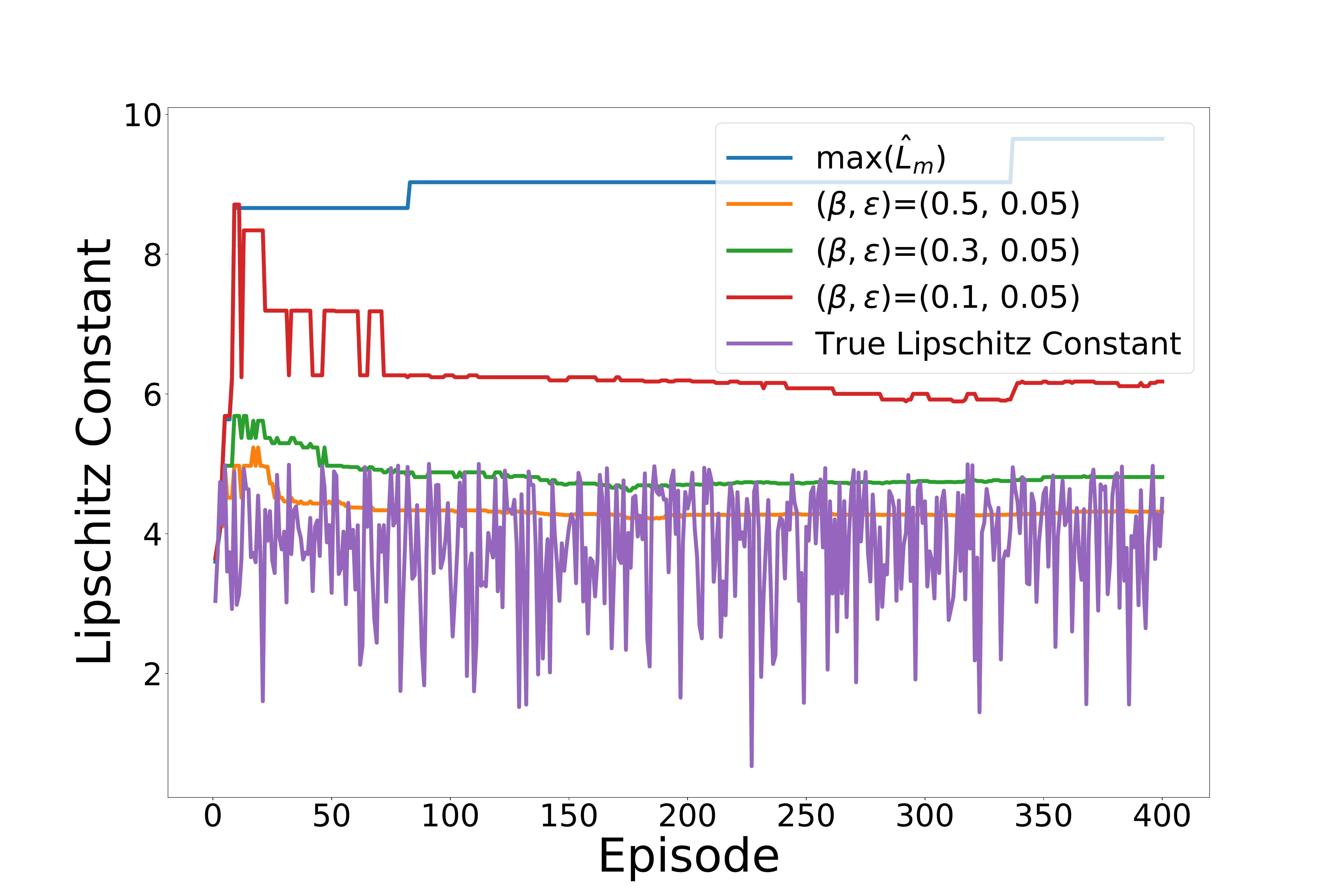}
        \label{fig:num-re:estimate:L5-2}
        }
    \subfigure[Cumulative regret over episodes
     ($L = 5$)]{
        \includegraphics[width=0.42\textwidth]{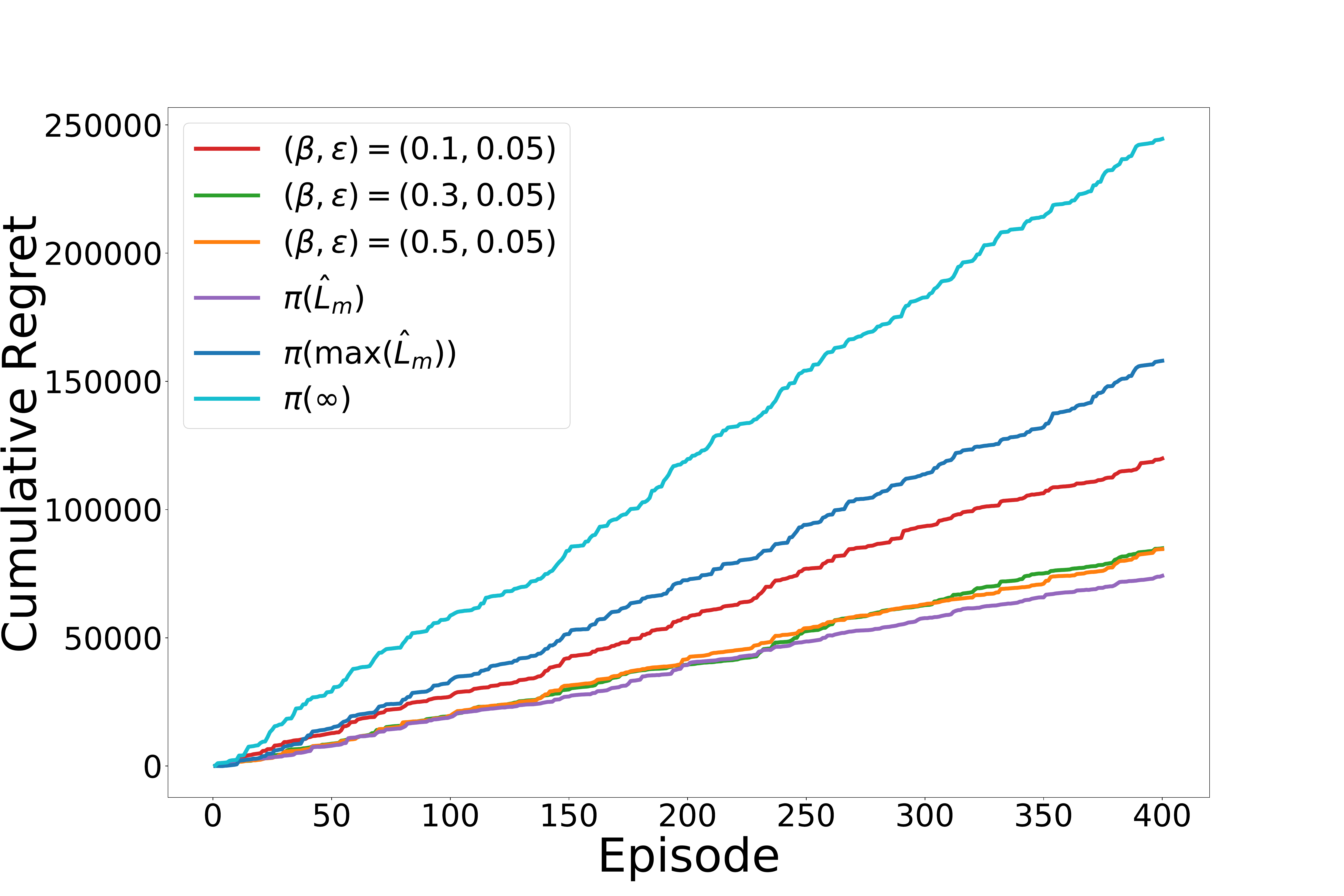}
        \label{fig:num-re:regret:L5-2}
        }
    \caption{Comparison of various estimators
     using the past experiences in Figure~\ref{fig:num:L}
     for $L  \in \{0.5, 5\}$} 
\end{figure*}



We present additional experiments to compare the performance of estimators on different Lipschitz constant $L \in \{0.5, 5\}$.

\noindent{\bf Setup.}
For numerical simulation, 
we consider Lipschitz structure $\Phi(L)$
with 
$L \in \{0.5, 5\}$,
$K = 6$ and $\vec{x} = (0, 0.8, 0.85, 0.9, 0.95, 1)$. 
For each episode $m \in [M= 400]$
of length $T = 10,000$, we generate $\vec{\mu}_m \in \Phi (L)$ as shown in Section~\ref{sec:numerical}.
For  each of $L = 0.5$ and $L = 5$,
Figures~\ref{fig:num:eg:L0.5}
and~\ref{fig:num:eg:L5-2}, respectively,
show five $\vec{\mu}_m$'s generated from the generative procedure,
and
Figures~\ref{fig:num:hist:L0.5}
and~\ref{fig:num:hist:L5}, respectively,
present the histogram of 400 $L_m$'s. For every estimator (stated below), for $L \in \{ 0.5, 5\}$, we use the same sequence of $\hat{L}_m$'s generated by $\pi(\infty)$ that uses no continuity structures.

\noindent{\bf Stable estimator $\hat{L}_\beta$.}
We compare four estimators on Lipschitz constant: 
three $\hat{L}_\beta$'s with $(\beta, \varepsilon_\beta) \in \{ (0.5, 0.05)$, $(0.3, 0.05)$, $(0.1, 0.05) \}$ and $\hat{L}_{\max}$ which takes 
the maximum of $\hat{L}_m$'s estimated previously.
The histogram of empirical $\hat{L}_m$ 
is presented in Figures~\ref{fig:num:hist:L0.5}~
and~\ref{fig:num:hist:L5}. 
Figures~\ref{fig:num-re:estimate:L0.5}~
and~\ref{fig:num-re:estimate:L5-2}
compare the accuracy of estimators.
As expected, for both $L = 0.5$ and $L = 5$, the most conservative estimation of  $\hat{L}_{\max}$ has monotonically increasing estimation of $L$
as the past episodes piling up. 
Theoretically, 
$\hat{L}_{\max}$ can explode up to $1/\Delta_{\vec{x}} =20$ in a finite number of episodes with positive probability. 
However, each of $\hat{L}_\beta$'s 
is stabilizing the estimation on Lischiptz constant $L$ regardless of its value 
once we collected sufficient experiences, e.g., $M > 30$.
Comparing the histograms of true $L_m$'s 
for different values of $L$ (Figures~\ref{fig:num:hist:L0.5}~and~\ref{fig:num:hist:L5}),
the portion of tight $L_m$ 
for $L =0.5$ is more than that for $L =5$. 
Hence, when $L =0.5$, we can obtain 
a safe estimation $\hat{L}_\beta \ge L$
even for the aggressive choice of $\beta = 0.5$.
However, when $L =5$, the same choice of $\beta$
gives a risky estimation $\hat{L}_\beta \lesssim L$,
in which padding some $\varepsilon_\beta$ may be helpful.

\noindent{\bf Cumulative regret.}
We present cumulative regret under various estimators 
in Figures~\ref{fig:num-re:regret:L0.5}~and~\ref{fig:num-re:regret:L5-2}.
Note that
the embedding of arm~1 is far from those of the other arms~$2$ to $6$. Hence, we anticipate that the observations among 
the neighboring arms~$2$ to $6$ are easily generalized to each other via Lipschitz continuity. In particular, the gain from Lipschitz continuity is much larger for $L=5$ than $L=0.5$
since the suboptimality gap $\Delta_{\vec{\mu}}$ 
can be larger for $L=5$ than $L=0.5$.
In Figure~\ref{fig:num-re:regret:L5-2}, 
as discussed in our analysis, more accurate estimation of $L$
consequences greater reduction in regret. 
Employing small value of $\beta$
provides much conservative estimation of $L$, 
but it can be too conservative to exploit the Lipschitz constant.

\end{document}